\documentclass[preprint,12pt]{elsarticle}

\usepackage{amssymb}
\usepackage{amsmath}
\usepackage{xspace}
\usepackage{paralist}
\usepackage{verbatim}
\usepackage{dsfont}
\usepackage{bbm}
\usepackage{todonotes}
\usepackage{graphicx}
\usepackage{hyperref}
\usepackage{subcaption}
\usepackage{booktabs}

\usepackage{tikz}
\usetikzlibrary{automata, positioning, fit, arrows, matrix} % <- attenzione a matrix!
\usepackage{multirow}

\newcommand{\ltl}{\ensuremath{\textsc{LTL}}\xspace}
\newcommand{\ltlf}{\ensuremath{\ltl_f}\xspace}

\DeclareMathOperator*{\argmax}{argmax}

\journal{Information Systems}

\begin{document}

\begin{frontmatter}



%\title{NeSyPPM: A Neuro-Symbolic Framework for Suffix Prediction in Predictive Process Monitoring \\OPPURE\\ Neuro-Symbolic Predictive Process Monitoring \\OPPURE\\ Neurosymbolic Autoregression for Predictive Process Monitoring: An Automata-Based Approach}

\title{Neuro-Symbolic Predictive Process Monitoring}


\author[unibz]{Axel Mezini\fnref{label1}}
\author[sapienza]{Elena Umili\fnref{label1}}
\author[unibz]{Ivan Donadello}
\author[unibz]{Fabrizio Maria Maggi}
\author[sapienza]{Matteo Mancanelli} 
\author[sapienza]{Fabio Patrizi} 
\fntext[label1]{Equal contribution} 

\affiliation[unibz]{organization={Faculty of Engineering, Free University of Bozen-Bolzano},
            addressline={NOI Techpark - via Bruno Buozzi, 1},
            city={Bolzano},
            postcode={39100}, 
            country={Italy}}

\affiliation[sapienza]{organization={Department of Computer, Control and Management Engineering, Sapienza, Università di Roma},
            addressline={Via Ariosto, 25}, 
            city={Roma},
            postcode={00185}, 
            country={Italy}}

\begin{abstract}
This paper addresses the problem of suffix prediction in Business Process Management (BPM) by proposing a Neuro-Symbolic Predictive Process Monitoring (PPM) approach that integrates data-driven learning with temporal logic-based prior knowledge. While recent approaches leverage deep learning models for suffix prediction, they often fail to satisfy even basic logical constraints due to the lack of explicit integration of domain knowledge during training. We propose a novel method to incorporate Linear Temporal Logic over finite traces (\ltlf) into the training process of autoregressive sequence predictors. Our approach introduces a differentiable logical loss function, defined using a soft approximation of \ltlf semantics and the Gumbel-Softmax trick, which can be combined with standard predictive losses. This ensures that the model learns to generate suffixes that are both accurate and logically consistent. Experimental evaluation on three real-world datasets shows that our method improves suffix prediction accuracy and compliance with temporal constraints. We also introduce two variants of the logic loss (local and global) and demonstrate their effectiveness under noisy and realistic settings. While developed in the context of BPM, our framework is applicable to any symbolic sequence generation task and contributes to advancing Neuro-Symbolic AI.
\end{abstract}


%\item Integration of linear temporal logic in neural training: two automata-based logical losses.
%\item Better logical constraints satisfaction of the automata-based logical losses.
%\item Important step towards green AI: an automata-based logical loss drastically reduces the training epochs.

%% Keywords
\begin{keyword}
Suffix prediction \sep Neuro-Symbolic AI \sep Deep learning with logical knowledge \sep Linear Temporal Logic \sep Differentiable automata

\MSC 68T05 \sep 68T37
\end{keyword}

\end{frontmatter}


\section{Introduction}
\label{sec:intro}

Predictive Process Monitoring (PPM)~\cite{Di_Francescomarino2026-uy} is a field within Business Process Management (BPM) that focuses on forecasting future events or outcomes of ongoing process instances based on historical event data. This paper addresses the PPM problem of suffix prediction by leveraging both data and prior logical temporal knowledge. Suffix prediction is particularly important in BPM for forecasting the continuation of a process trace, enabling better resource allocation, and anticipating future steps for effective decision-making.

Recently, there has been significant interest in employing deep learning techniques for suffix prediction in BPM~\cite{deeplearning_BPM_survey}, including the use of Recurrent Neural Networks (RNNs)~\cite{Tax17}, Transformers~\cite{transformer_BPM}, and Deep Reinforcement Learning algorithms~\cite{ramamaneiro24}. Despite these advances, several studies have shown that deep models can surprisingly fail to satisfy even the most basic logical constraints~\cite{roadR_giunchiglia, STLnet}, derived from commonsense reasoning or domain-specific knowledge. This occurs because such models are trained solely on data, and integrating logical knowledge into the training process remains an open challenge. As a result, in domains like BPM, where both data and formal knowledge about the process are often available, the latter is typically underutilized.

However, domain-specific knowledge can play a crucial role in characterizing the context in which a process is executed, providing valuable information not explicitly contained in the event data alone. For example, when the particular variant of a process being executed is known (such as a client-specific scenario or a seasonally influenced workflow), this knowledge can guide predictions in ways that purely data-driven models cannot. Similarly, in environments subject to concept drift, where process behavior evolves over time, logical constraints can help identify and adapt to changes more robustly than relying solely on historical data. Furthermore, logical knowledge is instrumental in filtering out noise in the data~\cite{SerafiniGBDSB21}, enhancing model robustness and prediction accuracy, as demonstrated in our experimental evaluation.

This work explores a novel Neuro-Symbolic PPM approach aimed at bridging this gap by incorporating prior knowledge, expressed in Linear Temporal Logic over finite traces (\ltlf), into the training of a deep generative model for suffix prediction. Existing works on knowledge-constrained sequence generation using autoregressive models are typically designed for test-time inference~\cite{DiFrancescomarino17, trident, llm_beam_search_1, llm_beam_search_2, llm_aux_mod_1, llm_aux_mod_2, llm_sampling1, montecarlo_llm, abs-2312-08847}, rather than for integration during training. A notable exception is STLnet~\cite{STLnet}, which incorporates Signal Temporal Logic (STL) specifications into the training process through a student-teacher framework. \textcolor{black}{However, STLnet is specifically designed for continuous-time sequences and STL constraints. Since STL generalizes LTL, we attempted to adapt STLnet to our symbolic domain by first translating our \ltlf formulas into STL and then applying STLnet. In practice, however, we found that the system was unable to construct the corresponding formula model without exceeding memory limits.\footnote{This is likely due to the fact that, in STL, temporal operators are applied over dense-time intervals (e.g., 
$G_{[1,2]}$ means “globally over all times in the interval $[1,2]$”), which are typically relatively short. In contrast, in \ltlf each operator ranges over the entire length of the trace. As a result, STL formulas obtained from \ltlf translations involve much larger temporal intervals, substantially complicating the application of the STLnet framework.} This highlights the limitations of STLnet in symbolic settings such as those encountered in PPM.}

Our contribution is a Neuro-Symbolic method to integrate symbolic knowledge of certain process properties, expressed in Linear Temporal Logic over finite traces (\ltlf), into the training process of a neural sequence predictor. This enables us to leverage both sources of information (data and background \ltlf knowledge) during training. Specifically, we achieve this by defining a differentiable counterpart of the \ltlf knowledge and employing a differentiable sampling technique known as the Gumbel-Softmax trick~\cite{gumb_sftmx_1}. By combining these two elements, we define a logical loss function that can be used alongside any other loss function employed by the predictor. This ensures that the network learns to generate traces that are both similar to those in the training dataset and compliant with the given temporal specifications \emph{at the same time}.

We evaluate our method on several real-world BPM datasets and demonstrate that incorporating \ltlf knowledge at training time leads to predicted suffixes with both lower Damerau-Levenshtein (DL) distance~\cite{DL_distance} from the target suffixes and a significantly higher rate of satisfaction of the \ltlf constraints. \textcolor{black}{This paper builds upon our previous work~\cite{UmiliPMAI24}, extending the approach to make it more principled, robust, and practically applicable in real-world scenarios. The main differences between the present work and~\cite{UmiliPMAI24} are:}

\begin{itemize}
\item \textcolor{black}{We provide a more rigorous formalization of the problem of autoregressive sequence generation under logical temporal constraints in Section~\ref{sec:problem_formulation}, which is entirely new. In the previous version, the problem setting and objectives were presented in a more concise and intuitive manner.}

\item \textcolor{black}{Building on this formalization, we offer a more exhaustive analysis of the differences in \emph{monitorability}~\cite{monitorability_LTL} across various types of temporal constraints when related to autoregressive sequence generation. This analysis leads us to introduce the notions of \textit{local} and \textit{global} guidance, described in Sections~\ref{sec:loc_guide} and~\ref{sec:glob_guide}, respectively.}

\item \textcolor{black}{Based on these two concepts, we formulate \textit{two} logical loss functions to guide the model toward satisfying the specification: one providing local, activity-level feedback, and another enforcing global, trace-level logical constraints during suffix prediction. The Local Logic Loss is entirely new, while the Global Logic Loss builds upon the loss proposed in~\cite{UmiliPMAI24} by introducing a Monte Carlo approximation phase. This addition both stabilizes learning and aligns the practical implementation of the loss with its theoretical formulation given in Section~\ref{sec:problem_formulation}.}

\item \textcolor{black}{We describe in more detail how we preprocess \(\ltlf\) knowledge to handle the sequence termination symbol (Section~\ref{sec:knowledge_preprocessing}), and how we transition from the automaton representation to its neural counterpart, DeepDFA~\cite{deepdfa_ecai2024}, which is used for loss computation (Section~\ref{sec:knowledge_tensorization}). Both of these components are new with respect to~\cite{UmiliPMAI24}.}

\item \textcolor{black}{Importantly, we substantially broaden the empirical evaluation of our framework, demonstrating both the effectiveness and generality of our approach. While~\cite{UmiliPMAI24} reported experiments only on synthetic datasets, here we present a comprehensive set of results on \textit{real-world} BPM datasets under challenging and \textit{noisy} experimental conditions. Consequently, Section~\ref{sec:exp} is also entirely new.}

\item \textcolor{black}{Finally, we provide a more detailed discussion of the potential applications of our framework beyond PPM, as well as the limitations that may arise in such broader settings, together with possible mitigation strategies in Section~\ref{sec:limitations}, which is also new.}
\end{itemize}

Overall, while our approach is instantiated in the context of PPM, its underlying principles are broadly applicable to any multi-step symbolic sequence generation task using autoregression. We believe the contributions of this work may be of general interest to the Machine Learning community, particularly in the area of Neuro-Symbolic AI.

The rest of the paper is structured as follows. Section~\ref{sec:bkg} provides preliminary notions and notations necessary to understand the paper. Section~\ref{sec:method} formulates the addressed problem and outlines the proposed solution, together with possible extensions and limitations. Section~\ref{sec:exp} presents an experimental evaluation of the solution. Section~\ref{sec:related} discusses related work, while Section~\ref{sec:conclusion} concludes the paper and spells out directions for future research.

% , including the broader community of researchers in Machine Learning, particularly those involved in neurosymbolic AI.

% In this particular work, we primarily refer to the sub-community of BPM researchers, due to the unique synergy between the use of temporal logics and the BPM field. 
\section{Background and Notation}
\label{sec:bkg}
This section introduces the fundamental notions and notations essential for understanding the concepts discussed in the paper. It lays the groundwork by defining key terms and formalizing the terminology used throughout the work.

\subsection{Notation}
In this work, we consider \textit{sequential} data of various types, including both symbolic and subsymbolic representations. Symbolic sequences are also called \textit{traces}. Each element in a trace is a symbol $\sigma$ drawn from a finite alphabet $\Sigma$. 
We denote sequences using bold notation. For example, $\boldsymbol{\sigma} = (\sigma_{1}, \sigma_{2}, \ldots, \sigma_{T})$ represents a trace of length $T$. 

Each symbolic variable in the sequence can be grounded either categorically or probabilistically.
In the case of categorical grounding, each element of the trace is assigned a symbol from $\Sigma$, denoted simply as $\sigma_{i}$, where $\sigma_{i}$ can be encoded as an index in $\{1, 2, \ldots, |\Sigma|\}$ or as a \textit{one-hot vector} $\sigma_{i} \in \{0,1\}^{|\Sigma|}$ such that $\sum_{j=1}^{|\Sigma|} \sigma_{i}[j] = 1$. In the case of probabilistic grounding, each symbolic variable is associated with a probability distribution over $\Sigma$, represented as a vector $\tilde{\sigma}_{i} \in \Delta(\Sigma)$, where $\Delta(\Sigma)$ denotes the probability simplex defined as:

\[
\Delta(\Sigma) = \left\{ \tilde{\sigma} \in \mathbb{R}^{|\Sigma|} \,\middle|\, \tilde{\sigma}[j] \geq 0,\ \sum_{j=1}^{|\Sigma|} \tilde{\sigma}[j] = 1 \right\}.
\]

Accordingly, we distinguish between categorically grounded sequences $\boldsymbol{\sigma}$ and probabilistically grounded sequences $\boldsymbol{\tilde{\sigma}}$, using the tilde notation.

Finally, we use subscripts to indicate time steps in the sequence, and $v[j]$ ($M[j]$) to denote the $j$-th component (tensor) of vector $v$ (matrix $M$). For instance, $\tilde{\sigma}_{i}[j]$ denotes the $j$-th component of the probabilistic grounding of $\sigma$ at time step $i$. 
We also use $+$ to denote trace concatenation; e.g., $\boldsymbol{a} + \boldsymbol{b}$ is the trace obtained by concatenating traces $\boldsymbol{a}$ and $\boldsymbol{b}$.

\subsection{Linear Temporal Logic and Deterministic Finite Automata} \label{sec:LTL_and_DFA}
Linear Temporal Logic (LTL) \cite{LTL} is a formal language that extends traditional propositional logic with modal operators, allowing the specification of rules that must hold \textit{through time}. In this work, we use \ltl interpreted over finite traces (\ltlf) \cite{LTLf}, which models finite, but length-unbounded, process executions, making it suitable for finite-horizon problems.

Given a finite set $\Sigma$ of atomic propositions, the set of \ltlf formulas $\phi$ is inductively defined as follows:
\begin{equation} \label{eq:LTL_syntax}
    \phi ::= \top \mid\bot\mid \sigma\mid\lnot\phi\mid\phi\land\phi\mid X\phi\mid\phi U\phi,
\end{equation}
where $\sigma \in \Sigma$. We use $\top$ and $\bot$ to denote $True$ and $False$, respectively. $X$ (Strong Next) and $U$ (Until) are temporal operators. Other temporal operators are $N$ (Weak Next) and $R$ (Release), defined as $N \phi \equiv \neg X\neg\phi$ and $\phi_1 R \phi_2 \equiv \neg(\neg\phi_1 U \neg\phi_2 )$; $G$ (Globally) $G\phi \equiv \bot R\phi$; and $F$ (Eventually) $F \phi \equiv \top U \phi$.

A trace $\boldsymbol{\sigma} = (\sigma_{1}, \sigma_{2}, \dots, \sigma_{T})$ is a sequence of propositional assignments to the propositions in $\Sigma$, where $\sigma_{t} \subseteq \Sigma$ is the set of all and only propositions that are true at instant $t$. Additionally, $|\boldsymbol{\sigma}| = T$ denotes the length of the trace. Since every trace is finite, $|\boldsymbol{\sigma}| < \infty$.
If the propositional symbols in $\Sigma$ are all \textit{mutually exclusive}, i.e., the domain produces exactly one symbol true at each step, then we have $\sigma_{t} \in \Sigma$. As is customary in BPM, we make this assumption, known as the \emph{mutual exclusivity assumption}~\cite{declare_assumptiopn}.
By $\boldsymbol{\sigma} \vDash \phi$ we denote that the trace $\boldsymbol{\sigma}$ satisfies the \ltlf formula $\phi$. 
We refer the reader to \cite{LTLf} for a formal description of the \ltlf semantics. 

Any \ltlf formula $\phi$ can be translated into a Deterministic Finite Automaton (DFA) \cite{LTLf} $A_\phi = (\Sigma, Q, q_0, \delta, F)$, where $\Sigma$ is the automaton alphabet,\footnote{Here the alphabet of the equivalent DFA is $\Sigma$ because of the mutual exclusivity assumption. In the general case, the automaton alphabet is $2^{\Sigma}$.} $Q$ is the finite set of states, $q_0 \in Q$ is the initial state, $\delta: Q \times \Sigma \rightarrow Q$ is the transition function, and $F \subseteq Q$ is the set of final states. 
Additionally, we recursively define the extended transition function over traces $\delta^*: Q \times \Sigma^* \rightarrow Q$ as:

\begin{equation}
\begin{array}{l}
    \delta^*(q,\epsilon) = q \\
    \delta^*(q, \sigma+\boldsymbol{x}) = \delta^*(\delta(q,\sigma) , \boldsymbol{x}),

\end{array}
\vspace{+0.3cm}
\end{equation}
where $\sigma \in \Sigma$ is a symbol and $\boldsymbol{x} \in \Sigma^*$ is a trace.  
The automaton accepts the trace $\boldsymbol{\sigma}$ if $\delta^*(q_0, \boldsymbol{\sigma}) \in F$, and in that case we say that $\boldsymbol{\sigma}$ belongs to the language of the automaton, denoted as $L(A_\phi)$.  
We have that $\phi$ and $A_\phi$ are equivalent because, for any trace $\boldsymbol{\sigma} \in \Sigma^*$:
\begin{equation}
 \boldsymbol{\sigma}\in L(A_\phi) \iff \boldsymbol{\sigma}\vDash\phi
 \vspace{-0.5cm}
\end{equation}
\textcolor{black}{\paragraph{Running Example}
We consider a domain with three possible activities \(\mathcal{A} = \Sigma= \{a,b,c\}\). For this domain, we are given both a set of traces describing example process executions and a constraint stating that activity \(a\) must always be eventually followed by activity \(b\). This constraint corresponds to the formula \(\phi = G(a \rightarrow F\,b)\), which is equivalent to the \textit{Response} constraint in Declare. The corresponding DFA, defined over the alphabet \(\Sigma = \{a,b\}\), is shown in Figure~\ref{fig:running_example} (Original DFA).}

\subsection{Deep Autoregressive Models and Suffix Prediction}
Deep autoregressive models are a class of deep learning models that automatically predict the next component in a sequence by using the previous elements in the sequence as inputs. These models can be applied to both continuous and categorical (symbolic) data, finding applications in various generative AI tasks such as Natural Language Processing (NLP) and Large Language Models (LLM) \cite{llama, gpt4}, image synthesis \cite{pixelRNN, PixelCNN}, and time-series prediction \cite{time_series_forecasting}. They encompass deep architectures such as RNNs and Transformers and, in general, any neural model capable of estimating the probability of a token given the preceding elements:

\begin{equation} \label{eq:next_activity_prob}
    P(x_{t} \mid x_{1}, x_{2}, \dots, x_{t-1}) = P(x_{t} \mid \boldsymbol{x_{<t}}).
\end{equation}

The probability of a sequence of data $\boldsymbol{x}$ can be calculated as:
\begin{equation} \label{eq:seq_prob}
    P(\boldsymbol{x}) = \prod_{i=1}^T P(x_{i} \mid \boldsymbol{x_{<i}}).
\end{equation}

In suffix prediction in BPM, given a subsequence (or prefix) of \textit{past activities} $\boldsymbol{p_t} = (a_{1}, a_{2}, \dots, a_{t})$ that the process has produced up to the current time step $t$, with $a_{i}$ in a finite set of activities $\mathcal{A}$, we aim to complete the trace by generating the sequence of future events, also called the \emph{suffix}, $\boldsymbol{s_t} = (a_{t+1}, \dots, a_{t+l}, \texttt{EOT})$.  
We use $\texttt{EOT} \notin \mathcal{A}$ to denote a special symbol marking the end of sequences, which is not included in $\mathcal{A}$.

Suffix prediction can be accomplished using autoregressive models, by choosing at each step the most probable next event according to the neural network, concatenating it with the prefix, and continuing to predict the next event in this manner until the \texttt{EOT} symbol is predicted or the trace has reached a maximum number of steps $T$. We define $\mathcal{A}^{\leq T}\texttt{EOT} \equiv \{\boldsymbol{a}+\texttt{EOT} \mid \boldsymbol{a} \in \mathcal{A}^{\leq T}\}$ as the set of possible complete traces $\boldsymbol{a} = \boldsymbol{p_t}+\boldsymbol{s_t}$ that can be generated in this way.

At each generation step, a symbol must be \textit{sampled} from the next activity probability. A common way of selecting the next activity to feed into the autoregressor is to \textit{greedily} choose the activity maximizing the next symbol probability at each step, as follows:

\begin{equation}
    \label{eq:suffix-generation}
    {a}_{k} = \argmax_{a \in \mathcal{A}\cup\{\texttt{EOT}\}} P(a_{t}= \sigma \mid a_{1}, \dots, a_{t}, {a}_{t+1}, \dots, {a}_{k-1}), \quad t < k \leq T.
\end{equation}

This greedy search strategy may not produce the \textit{most probable suffix}, i.e., the trace that maximizes the probability in Equation~\ref{eq:seq_prob}. Other non-optimal sampling strategies commonly used for this task include Beam Search, Random Sampling, and Temperature Sampling~\cite{ramamaneiro24}.
%The performance of suffix prediction strictly depends on both the neural model's ability to estimate the next event probability and the sampling strategy used. In this work, we employ RNNs to model the next event probability and use a Temperature Sampling strategy at test time, which is a trade-off between greedy and random search techniques. Nonetheless, the method we propose is general enough to be employed with any other deep autoregressive model and sampling strategy.

\section{Method}
\label{sec:method}
This section defines the specific suffix prediction problem we aim to solve and outlines our Neuro-Symbolic PPM approach, which integrates prior knowledge expressed in Linear Temporal Logic over finite traces (\ltlf) into the training of a deep generative model.
We also introduce two logic loss formulations: one providing \emph{local}, activity-level feedback, and another enforcing \emph{global}, trace-level constraints over the predicted suffix.

\subsection{Problem Formulation} \label{sec:problem_formulation}
We assume an autoregressive neural model $f_\theta$ with trainable parameters $\theta$ that estimates an approximation $P_\theta$ of the probability of the next event $a_t$ given a trace of previous events $\boldsymbol{a}_{<t}$ (Equation~\ref{eq:next_activity_prob}):
\begin{equation} \label{eq:RNN}
\begin{array}{ll}
     \tilde{y}_t = f_\theta(\boldsymbol{a}_{<t}) \\
     P(a_{t} = a_i \in \mathcal{A} \cup \{\texttt{EOT}\} \mid \boldsymbol{a}_{<t}) \approx \tilde{y}_t[i].
\end{array}
\end{equation}
Note that we do not make any assumptions about the neural model, except that it can estimate the probability of the next activity given a sequence of previous ones. As a result, our approach is entirely \textit{model-agnostic} and can be readily applied to any autoregressive model.

We denote by $P_\theta$ the probability of a trace according to the network approximation:
\begin{equation}
    P_\theta(\boldsymbol{a}) = \prod_{t=1}^{|\boldsymbol{a}|} \tilde{y}_t[a_t].
\end{equation}

The model parameters are typically trained using a supervised loss $L_{\mathcal{D}}$, evaluated on a dataset $\mathcal{D}$ of ground-truth traces obtained by observing the process. The loss for a trace $\boldsymbol{a} \in \mathcal{D}$ of length $T$ is defined as follows:
\begin{equation} \label{eq:sup_loss}
    L_{\mathcal{D}}(\boldsymbol{a}) = \frac{1}{T} \sum_{t=1}^T \text{cross-entropy}(f_\theta(\boldsymbol{a}_{< t}), a_{t}) .
\end{equation}
This loss trains the network to predict the next symbol in a trace so as to closely mimic the data in the dataset.

In this work, we assume that certain properties of the process are also known and can be injected into the learning process during training. These properties, which form the \emph{background} (or \emph{prior}) knowledge about the process, are expressed as an \ltlf formula $\phi$ defined over an alphabet $\Sigma \subseteq \mathcal{A}$.

Our goal is for the language generated by the autoregressor $f_\theta$ to be strictly contained within the language of strings accepted by the formula, denoted as $L(A_{\phi})$. However, the language produced by the network is \emph{unbounded}, as it is only \emph{softly assigned}: in other words, the network can generate any possible string, each with a different probability.

Our method therefore aims to maximize the probability $P_{\theta \vDash \phi}$ that traces $\boldsymbol{a} \sim P_\theta$, sampled from the autoregressor, satisfy the specification:
\begin{equation} \label{eq:P_theta_satisfy_phi}
    P_{\theta \vDash \phi} = \mathbb{E}_{\boldsymbol{a} \sim P_\theta}[\boldsymbol{a} \vDash \phi] = \sum_{\boldsymbol{a} \in \mathcal{A}^{\leq T}\texttt{EOT}} P_\theta(\boldsymbol{a}) \, \mathds{1}\{\boldsymbol{a} \vDash \phi\}.
\end{equation}
Note that, in order to compute the exact probability of knowledge satisfaction, one would need to enumerate \textit{all possible suffixes} of maximum length $T$ and sum their acceptance probabilities. However, this set has exponential size, making exact computation unfeasible. Furthermore, we aim to impose the maximization of this probability as a \textit{training objective}. Therefore, our goal is to design a fast and differentiable procedure to approximate it at each optimization step of the autoregressor.

We propose two logic loss functions: a local, activity-level loss $L_\phi^{\text{loc}}$, and a global, trace-level loss $L_\phi^{\text{glob}}$. We show that both contribute positively to the learning process, demonstrating the effectiveness of integrating \ltlf knowledge into autoregressive training. The choice between the two logic losses depends on the type of \ltlf knowledge available. The global loss $L_\phi^{\text{glob}}$ can always be applied, regardless of the structure of the formula $\phi$, but it is more computationally demanding, as it requires generating entire suffixes during training.

In general, the compliance of a trace with $\phi$ can only be assessed on complete traces. Indeed, whether the current partial prediction $\boldsymbol{a}^{\leq t}$ satisfies (or violates) $\phi$ does not guarantee that the final predicted trace $\boldsymbol{a}$ will also satisfy (or violate) it. This introduces a challenge in evaluating \ltlf constraints, as opposed to approaches that rely on local constraints~\cite{montecarlo_llm}, which can be verified step-by-step during generation.

However, some types of formulas, such as safety constraints, once translated into a DFA, contain \emph{failing states}, which can be used to guide generation locally. A failing state is a state in the DFA from which no accepting state is reachable. When a partial trace enters such a state, it has irreversibly violated the formula, and no continuation can satisfy the knowledge. Our local loss $L_\phi^{\text{loc}}$ exploits the presence of failing states to guide the autoregressor at every generation step. As a result, it provides richer feedback at the activity level, but it can only be used with formulas that admit such a DFA structure. \textcolor{black}{Nevertheless, this limitation is overcome during the DFA preprocessing phase, where a failing state is always added to handle the \texttt{EOT} symbol, as discussed in Section~\ref{sec:knowledge_preprocessing}.}

In the following sections, we describe how we compute the logic loss under the two scenarios. In either case, we combine it with the supervised loss $L_{\mathcal{D}}$ as follows:
\begin{equation}
\label{eq:loss-balance}
    L = \alpha L_{\mathcal{D}} + (1 - \alpha) L_\phi,
\end{equation}
with $\alpha$ being a constant between 0 and 1 that balances the influence of each loss on the training process.

\subsection{Local Guidance} \label{sec:loc_guide}

In cases where the formula can be permanently violated, the corresponding DFA includes a set of failing states $Q^{\text{fail}} \subseteq Q$, from which the formula can no longer be satisfied. In this case, we aim to minimize the \textit{probability that the next predicted activity will irreversibly violate prior knowledge}, denoted as $P(\boldsymbol{a}_t \nvDash^{\times} \phi \mid \boldsymbol{a}_{<t})$.
Here, the symbol $\nvDash^\times$ denotes the permanent violation of the formula.
Given a ground-truth trace $\boldsymbol{a} \in \mathcal{D}$, we define this probability as:
\begin{equation} \label{eq:irreversible_violation_prob}
  P(\boldsymbol{a}_t \nvDash^{\times} \phi \mid \boldsymbol{a}_{<t}) = \sum_{a \in \mathcal{A} \cup \{\texttt{EOT}\}} f_\theta(\boldsymbol{a}_{<t})[a] \cdot \mathds{1}\left\{\delta^*(q_0, \boldsymbol{a}_{<t} + a) \in Q^{\text{fail}}\right\}.
\end{equation}

In other words, at each step $t$ of the trace, we consider the probability distribution over the next symbol given by $f_\theta(\boldsymbol{a}_{<t})$. For each symbol $a$ in the vocabulary (including \texttt{EOT}), we simulate the state reached by the extended trace $\boldsymbol{a}_{<t} + a$ in the DFA. If the resulting state belongs to $Q^{\text{fail}}$, we add the probability of $a$ to the probability that the trace permanently violates the knowledge.
We minimize this probability by minimizing the following loss, hereafter called the \emph{Local Logic Loss} (LLL), on each ground-truth trace $\boldsymbol{a}$:
\begin{equation}
\label{eq:lll}
    L_\phi^{\text{loc}} = \frac{1}{|\boldsymbol{a}|} \sum_{t=1}^{|\boldsymbol{a}|} -\log\left(1 - P(\boldsymbol{a}_t \nvDash^{\times} \phi \mid \boldsymbol{a}_{<t})\right).
\end{equation}

Note that traces can fall into one of the following categories:
(i) they irreversibly violate the knowledge (by reaching a failing state either at termination or earlier);
(ii) they do not satisfy the knowledge (by terminating in a non-accepting, non-failing state);
(iii) they satisfy the knowledge (by terminating in an accepting state).

Both cases (i) and (ii) result in a violation of the knowledge. The loss $L_\phi^{\text{loc}}$ specifically targets only case (i), aiming to reduce occurrences of irreversible violations. By decreasing the probability in Equation~\ref{eq:irreversible_violation_prob}, we increase the desired probability $P_{\theta \vDash \phi}$. However, knowledge satisfaction is not directly maximized, and it is still possible for traces with zero loss to violate the knowledge under case (ii).
Despite this limitation, we observed in practice that $L_\phi^{\text{loc}}$ remains highly effective on many datasets, as we will show in Section~\ref{sec:exp}.

\subsection{Global Guidance} \label{sec:glob_guide}
In the general case, an \ltlf formula $\phi$ cannot always be permanently violated. In such cases, it is impossible to supervise the suffix generation step-by-step, and the only guidance that can be provided to the autoregressor is \textit{global}, i.e., applied to the entire generated trace.
To this end, we define a second logic loss (hereafter called the \emph{Global Logic Loss} (GLL)), $L_\phi^{\text{glob}}$, which provides this global supervision to the autoregressor and can be applied to \emph{any} \ltlf formula.

In particular, in the global case, we directly approximate the target probability $P_{\theta \vDash \phi}$ using a Monte Carlo estimation. We sample a set of complete traces $\{\boldsymbol{a}^{(1)}, \boldsymbol{a}^{(2)}, \ldots, \boldsymbol{a}^{(N)}\} \sim P_\theta$ according to the distribution learned by the autoregressor, and compute an approximation of the target probability $\hat{P}_{\theta \vDash \phi}$ as the empirical average compliance with the formula over the sampled set:
\begin{equation} \label{eq:montecarlo_approx}
    \hat{P}_{\theta \vDash \phi} = \frac{1}{N} \sum_{i=1}^N \mathds{1} \{ \boldsymbol{a}^{(i)} \vDash \phi \}.
\end{equation}
\textcolor{black}{However, the sampled traces $\{\boldsymbol{a}^{(1)}, \boldsymbol{a}^{(2)}, \ldots, \boldsymbol{a}^{(N)}\}$ used for the Monte Carlo estimation are generated by a neural network that returns a probability distribution over activity names at each time step. Therefore, a crisp trace $\boldsymbol{a}^{(i)}$ is replaced by its probabilistic counterpart $\tilde{\boldsymbol{a}}^{(i)}$, where activity symbols are sampled from the probability distribution computed by the neural network. Moreover, the indicator function $\mathds{1}$ in Equation~\ref{eq:montecarlo_approx} is replaced by a method that computes the (probabilistic) compliance of a probabilistic trace with the knowledge in a fast and differentiable way.}
To achieve this, our method relies on two key components:
\begin{enumerate}
    \item Gumbel-Softmax sampling~\cite{gumb_sftmx_1, gumb_sftmx_2} to generate differentiable, near one-hot suffixes during training. \textcolor{black}{A probabilistic trace $\tilde{\boldsymbol{a}}^{(i)}$ is therefore obtained by concatenating an input prefix with the sampled suffix.}
    \item DeepDFA~\cite{deepdfa_ecai2024}, a Neuro-Symbolic framework that encodes temporal logic properties as a recurrent layer, enabling efficient and differentiable evaluation of logical constraints. \textcolor{black}{This allows us to compute the probabilistic compliance $P_{\mathrm{DDFA}}(\tilde{\boldsymbol{a}}^{(i)} \vDash \phi)$ of a sampled trace $\tilde{\boldsymbol{a}}^{(i)}$ with the knowledge $\phi$. Therefore, Equation~\ref{eq:montecarlo_approx} becomes:
    \begin{equation} \label{eq:montecarlo_approx_ddfa}
    \hat{P}_{\theta \vDash \phi} = \frac{1}{N} \sum_{i=1}^N P_{\mathrm{DDFA}}(\tilde{\boldsymbol{a}}^{(i)} \vDash \phi).
\end{equation}}
\end{enumerate}
By leveraging these two components, detailed in the following sections, we compute the global logic loss $L_\phi^{\text{glob}}$, which enforces the satisfaction of prior knowledge over entire traces:
\begin{equation}
    L_\phi^{\text{glob}} = - \log(\hat{P}_{\theta \vDash \phi}).
\end{equation}

In the next sections, we first discuss further properties of the local and global guidance. Then, we describe how the \ltlf formula $\phi$ is preprocessed and encoded using DeepDFA. Finally, we illustrate how DeepDFA is integrated with suffix prediction during training through differentiable sampling based on Gumbel-Softmax.

\textcolor{black}{
\subsection{Further Discussions about Local and Global Guidance} 
The local and global guidance can be further compared along two dimensions: the training procedure and the scope of the $\ltlf$ properties that are exploited.} 

\textcolor{black}{During the training procedure, both types of guidance learn a similar conditional distribution, namely the probability that a trace is compliant with the knowledge given the context (i.e., the prefix), but they differ in how such a context is conditioned during training. The local guidance adopts the \emph{teacher forcing} training procedure, which trains the neural network model by feeding the true previous tokens $\boldsymbol{a}_{<t}$ as inputs at every step and minimizing the negative log-likelihood in Equation~\ref{eq:lll}. In general, this procedure yields stable and low-variance gradients. On the other hand, the global guidance exploits \emph{autoregressive training}, which feeds the neural network model’s own predictions $\tilde{\boldsymbol{a}}_t^{(i)}$ back as inputs. In this way, the loss becomes an expectation over model-generated prefixes, making optimization noisier and more sensitive to early mistakes. In general, teacher forcing is preferred for efficiency and stability, while autoregressive training better matches inference but is harder to optimize. The Monte Carlo estimation through sampling in GLL is performed to obtain a more stable autoregressive training procedure.}

\textcolor{black}{Regarding the scope of $\ltlf$ properties, Declare constraints are characterized by two well-known $\ltlf$ properties: safety and liveness. Safety constraints express that a certain behavior must never occur, whereas liveness constraints require that a condition will eventually be satisfied. When a safety constraint is translated into a deterministic finite automaton, a dedicated failing state is always generated to capture violations. In contrast, this is not the case for liveness constraints, since a trace can only be considered violated at its termination if the required condition has never been fulfilled. Figure~\ref{fig:dfa_absence_existence} illustrates the DFAs for the Absence (i.e., a certain activity cannot occur) and Existence (i.e., a certain activity must occur) Declare constraints, which exhibit the safety and liveness properties, respectively.}

\begin{figure}[t!]
\centering
\begin{tikzpicture}[auto, node distance=2.2cm, scale=0.9]
\color{black}

\node at (0, 4.1) {};

\begin{scope}[xshift=1.2cm]
  \node at (1.5, 2.8) {Existence(a)};
  \node[state, initial] (E0) at (0, 1.2) {$q_0$};
  \node[state, accepting] (E1) at (3, 1.2) {$q_1$};

  \path[->]
    (E0) edge[loop above] node {} ()
         edge[bend left=15] node {a} (E1)
    (E1) edge[loop above] node {} ();
\end{scope}

\begin{scope}[xshift=-5.2cm]
  \node at (1.5, 2.8) {Absence(a)};
  \node[state, initial, accepting] (A0) at (0, 1.2) {$q_0$};
  \node[state] (A1) at (3, 1.2) {$q_1$};

  \path[->]
    (A0) edge[loop above] node {} ()
         edge[bend left=15] node {a} (A1)
    (A1) edge[loop above] node {} ();
\end{scope}

\begin{comment}
    
\node at (0, -0.4) {\textbf{DFA after preprocessing}};

\begin{scope}[xshift=1.2cm, yshift=-4.4cm]
  \node at (1.5, 2.8) {Existence(a)};
  \node[state, initial] (e0) at (0.4, 1.2) {$q_0$};
  \node[state] (e1) at (2.6, 1.2) {$q_1$};
  \node[state, accepting] (e2) at (2.6, -1) {$q_2$};
  \node[state] (e3) at (0.4, -1) {$q_3$};           

  \path[->]
    (e0) edge[loop above] node {b,c} ()
         edge[bend left=15] node {a} (e1)
    (e1) edge[loop above] node {b,c} ()
    (e1) edge node[right] {\texttt{EOT}} (e2)
    (e0) edge node[left]  {\texttt{EOT}} (e3)
    (e2) edge[loop below] node {b} ()
    (e3) edge[loop below] node {b} ();
\end{scope}

\begin{scope}[xshift=-5.2cm, yshift=-4.4cm]
  \node at (1.5, 2.8) {Absence(a)};
  \node[state, initial] (a0) at (0.4, 1.2) {$q_0$};
  \node[state] (a1) at (2.6, 1.2) {$q_1$};
  \node[state, accepting] (a2) at (0.4, -1) {$q_2$}; 
  \node[state] (a3) at (2.6, -1) {$q_3$};           

  \path[->]
    (a0) edge[loop above] node {} ()
         edge[bend left=15] node {a} (a1)
    (a1) edge[loop above] node {} ()
    (a0) edge node[left]  {\texttt{EOT}} (a2)
    (a1) edge node[right] {\texttt{EOT}} (a3)
    (a2) edge[loop below] node {} ()
    (a3) edge[loop below] node {} ();
\end{scope}
\end{comment}

\end{tikzpicture}
\caption{\textcolor{black}{DFAs obtained after the conversion of the Absence (on the left) and Existence (on the right) constraints for activity $a$. The Absence DFA has a failing state, $q_1$, from which no accepting states can be reached (safety property). The Existence DFA has no failing states; from all states, an accepting state can be reached (liveness property).}}

\label{fig:dfa_absence_existence}
\end{figure}
\textcolor{black}{The Absence constraint exhibits the safety property, whereas the Existence constraint exhibits the liveness property. The local guidance leverages the presence of failing states by penalizing the model whenever it assigns a high probability to transitions leading to such states. This mechanism enables the model to receive immediate feedback whenever a violation occurs and, therefore, is best suited for constraints that exhibit the safety property. Conversely, liveness constraints do not inherently produce explicit failing states. The only exploitable failing state is therefore the one introduced during the preprocessing phase (see Section~\ref{sec:knowledge_preprocessing}), which can be reached exclusively through the end-of-trace signal. As a result, in this case, the violation signal can only be provided at the conclusion of the trace, similarly to the behavior of the global guidance.}

\subsection{Knowledge Preprocessing and Tensorization}
In this section, we describe how the \ltlf formula $\phi$ is preprocessed and \emph{tensorized} to enable the integration of prior knowledge into the learning process. Note that these steps are performed only once, before training begins, and are required to compute both the local and global losses.

\begin{figure}[t!]
\centering
\begin{tikzpicture}[scale=0.85, every node/.style={scale=0.85}]
\color{black}

\begin{scope}[xshift=-6cm, yshift=3cm]
\node at (1.5,3.5) {\textbf{Original DFA $A_\phi$}};

\node[state, initial, accepting] (q0) at (0,1) {$q_0$};
\node[state] (q1) at (3,1) {$q_1$};

\path[->]
  (q0) edge[loop above] node {b} ()
       edge[bend left=15] node[pos=0.5, yshift=-2mm] {a} (q1)
  (q1) edge[loop above] node {a} ()
       edge[bend left=15] node[pos=0.5, yshift=-2mm] {b} (q0);
\end{scope}

\begin{scope}[xshift=-6cm, yshift=-1cm]
\node at (1.5,3.5) {\textbf{Preprocessed DFA $A_\phi'$}};

\node[state, initial] (p0) at (0,1) {$q_0$};
\node[state] (p1) at (3,1) {$q_1$};
\node[state, accepting] (p2) at (0,-1) {$q_2$};
\node[state] (p3) at (3,-1) {$q_3$};

\path[->]
  (p0) edge[loop above] node {b,c} ()
       edge[bend left=15] node[pos=0.5, yshift=-2mm] {a} (p1)
       edge node[left] {\texttt{EOT}} (p2)
  (p1) edge[loop above] node {a,c} ()
       edge[bend left=15] node[pos=0.5, yshift=-2mm] {b} (p0)
       edge node[right] {\texttt{EOT}} (p3)
  (p2) edge[loop below] node {a,b,c,\texttt{EOT}} ()
  (p3) edge[loop below] node {a,b,c,\texttt{EOT}} ();
\end{scope}

% ---------------- Tensorization (right panel) ----------------
\begin{scope}[xshift=4cm, yshift=2.5cm]
\node at (0,3.5) {\textbf{Tensorization of DFA $A_\phi'$}};

% Initial and final state vectors side by side
\node at (-2.0,2.5) {\textbf{Initial state $\mu$}};
\node at (-2.0,1.5) {$\Big($\ \textbf{1} 0 0 0\ $\Big)$};

\node at (2,2.5) {\textbf{Final states $\lambda$}};
\node at (2,1.5) {$\Big($\ 0 0 \textbf{1} 0\ $\Big)$};

% Transition matrices 2x2 (T[a], T[b] sopra; T[c], T[EOT] sotto)
\node at (-2,0.5) {$\mathcal{T}[a]$};
\node at (-2,-1) {$\begin{pmatrix}
0 & \textbf{1} & 0 & 0 \\
0 & \textbf{1} & 0 & 0 \\
0 & 0 & \textbf{1} & 0 \\
0 & 0 & 0 & \textbf{1}
\end{pmatrix}$};

\node at (2,0.5) {$\mathcal{T}[b]$};
\node at (2,-1) {$\begin{pmatrix}
\textbf{1} & 0 & 0 & 0 \\
\textbf{1} & 0 & 0 & 0 \\
0 & 0 & \textbf{1} & 0 \\
0 & 0 & 0 & \textbf{1}
\end{pmatrix}$};

\node at (-2,-3.0) {$\mathcal{T}[c]$};
\node at (-2,-4.5) {$\begin{pmatrix}
\textbf{1} & 0 & 0 & 0 \\
0 & \textbf{1} & 0 & 0 \\
0 & 0 & \textbf{1} & 0 \\
0 & 0 & 0 & \textbf{1}
\end{pmatrix}$};

\node at (2,-3.0) {$\mathcal{T}[\texttt{EOT}]$};
\node at (2,-4.5) {$\begin{pmatrix}
0 & 0 & \textbf{1} & 0 \\
0 & 0 & 0 & \textbf{1} \\
0 & 0 & \textbf{1} & 0 \\
0 & 0 & 0 & \textbf{1}
\end{pmatrix}$};

\end{scope}

\end{tikzpicture}
\caption{Preprocessing and tensorization for the running example. Left: original DFA $A_\phi$ (top) and preprocessed DFA $A_\phi'$ (bottom). Right: tensorization of $A_\phi'$ with initial state vector $\mu$, final state vector $\lambda$, and transition matrices $\mathcal{T}[a], \mathcal{T}[b], \mathcal{T}[c], \mathcal{T}[\texttt{EOT}]$, where a 1 in position $(i,j)$ of $\mathcal{T}[x]$ indicates a transition $q_i \xrightarrow{x} q_j$ in the automaton.}
\label{fig:running_example}
\end{figure}

\subsubsection{Knowledge Preprocessing} \label{sec:knowledge_preprocessing}

%1 traduzione da LTLf a automa
First, we translate the \ltlf formula $\phi$ into an equivalent deterministic finite automaton (DFA) $A_\phi = (\Sigma, Q, q_0, \delta, F)$ using the automatic translation tool \texttt{ltlf2DFA}~\cite{fuggitti-ltlf2dfa}. This step is necessary because all existing approaches for integrating temporal specifications into learning pipelines rely on automata-based representations~\cite{deepdfa_ecai2024, umili_kr23, NesyA}, while the direct integration of \ltlf formulas remains an open challenge~\cite{fuzzyLog_donadello, DonadelloFIMM25}.
Although this translation has worst-case \emph{double-exponential} complexity~\cite{LTLf}, it is often efficient in practice, and several scalable techniques exist to perform it for a given formula~$\phi$~\cite{LTL2DFA1, LTL2DFA2, LTL2DFA3}. Moreover, in this work, we focus on \textit{Declare} formulas~\cite{declare2006}, a standard for declaratively specifying business processes~\cite{declare2007}, which are known to yield DFAs of \emph{polynomial size} with respect to the input formula~\cite{declare2011}.

Second, we adapt the DFA alphabet $\Sigma \subseteq \mathcal{A}$ so that it includes all activity symbols present in the event log $\mathcal{A}$. Specifically, for each symbol $s \in \mathcal{A} \setminus \Sigma$ and each state $q \in Q$, we add a self-loop $\delta(q, s) = q$ to the transition function $\delta$. This ensures that symbols not constrained by the formula can still be processed by the DFA without affecting its acceptance behavior.
Note that while this technique is feasible in the BPM setting, it may become impractical in other application domains where the autoregressor’s symbol space is excessively large, such as in LLM-based applications~\cite{trident}.

Additionally, we extend the DFA to handle the special \texttt{EOT} (End-of-Trace) symbol. In particular, we define as accepted all and only those traces of the form $t{+}\texttt{EOT}{+}z$ such that $t \in \mathcal{A}^*$, $t \vDash \phi$, and $z$ is any (possibly empty) trace in $(\mathcal{A} \cup \{\texttt{EOT}\})^*$.
This implies that:
\begin{itemize}
    \item[(i)] a non-terminating trace is considered \emph{non-compliant} with the specification;
    \item[(ii)] a trace is evaluated \emph{against the specification only at the first occurrence of \texttt{EOT}} -- whether the specification is violated or satisfied \emph{before} this point is irrelevant, as are any symbols in $z$ occurring after the first \texttt{EOT}.
\end{itemize}

To implement this behavior, we add \texttt{EOT} to the alphabet and introduce two terminal states: a success state $q_t^s$ and a failure state $q_t^f$. For each state $q \in Q$, we add the transition $\delta(q, \texttt{EOT}) = q_t^s$ if $q \in F$, and $\delta(q, \texttt{EOT}) = q_t^f$ otherwise.
States $q_t^s$ and $q_t^f$ are \emph{terminal}, meaning they are absorbing: once entered, the automaton cannot exit. Therefore, for each symbol $s \in \mathcal{A} \cup \{\texttt{EOT}\}$, we add the transitions $\delta(q_t^s, s) = q_t^s$ and $\delta(q_t^f, s) = q_t^f$.
We designate $q_t^s$ as the \emph{only accepting state} of the extended DFA.

Finally, we compute the set of failure states $Q^{\text{fail}}$, which is only necessary when employing the local logic loss $L_\phi^{\text{loc}}$.
The final DFA after preprocessing is therefore:
\[
A_\phi' = (\mathcal{A} \cup \{\texttt{EOT}\}, Q \cup \{q_t^s, q_t^f\}, q_0, \delta', \{q_t^s\}),
\]
with the extended transition function $\delta'$ defined as:
\begin{equation}
    \delta'(q,s) =
\begin{cases}
\delta(q,s) & \text{if } q \in Q \wedge s \in \Sigma, \\
q & \text{if } s \in \mathcal{A} \setminus \Sigma \vee q \in \{q_t^s, q_t^f\}, \\
q_t^s & \text{if } s = \texttt{EOT} \wedge q \in F, \\
q_t^f & \text{if } s = \texttt{EOT} \wedge q \notin F.
\end{cases}
\end{equation}

\textcolor{black}{
Note that the input background knowledge could contain conflicting constraints, meaning that their conjunction cannot be satisfied. Therefore, the corresponding automaton is empty, with no accepting states. In this case, our system is still able to learn, but the contribution of the conflicting input knowledge is ignored. Indeed, both logical losses have a constant value that cannot be reduced by updating the neural network weights. Consequently, only the task loss is able to drive the learning.
\paragraph{Running Example: DFA Preprocessing}
The original DFA shown in Figure~\ref{fig:running_example} is transformed into $A_\phi'$, illustrated in the figure. Note that the new DFA is defined over all symbols in the domain plus the termination symbol, i.e., $\Sigma' = \mathcal{A} \cup \{\texttt{EOT}\}$, and contains the same states as $A_\phi$ together with two terminal states: one accepting terminal state ($q_t^s = q_2$) and one failing terminal state ($q_t^f = q_3$).}

\subsubsection{Knowledge Tensorization} \label{sec:knowledge_tensorization}
Given the final DFA $A_\phi'$, we transform it into a neural layer using DeepDFA~\cite{deepdfa_ecai2024}. DeepDFA is a neural, probabilistic relaxation of a standard deterministic finite-state machine, where the automaton is represented in matrix form and the input symbols, states, and outputs are \textit{probabilistically grounded}.

We define the transition function of the DFA in matrix form as $\mathcal{T} \in \mathbb{R}^{|\Sigma| \times |Q| \times |Q|}$, and the initial and final states as the vectors $\mu \in \mathbb{R}^{|Q|}$ and $\lambda \in \mathbb{R}^{|Q|}$, respectively.
While this matrix representation is traditionally used for Probabilistic Finite Automata (PFA), here it is applied to deterministic automata to leverage tensor operations for fast and differentiable computation of state transitions and outputs.
The DeepDFA model is defined as follows:
\begin{equation} \label{eq:deepDFA}
\begin{aligned}
    \tilde{q}_{0} &= \mu, \\
    \tilde{q}_{t} &= \sum_{j=1}^{|\Sigma|} \tilde{\sigma}_t[j] \cdot (\tilde{q}_{t-1} \cdot \mathcal{T}[j]), \\
    \tilde{o}_{t} &= \tilde{q}_{t} \cdot \lambda^\top.
\end{aligned}
\end{equation}
Here, $\tilde{\sigma}_t$, $\tilde{q}_t$, and $\tilde{o}_t$ represent the probabilistic representations of the input symbol, the automaton state, and the output (i.e., whether the trace is accepted or rejected) at time $t$. \textcolor{black}{In the GLL, the probabilistically grounded sequence $\tilde{\boldsymbol{\sigma}}$ is the sampled trace $\tilde{\boldsymbol{a}}^{(i)}$; therefore, the output $\tilde{o}_t$ corresponds to the probability $P_{\mathrm{DDFA}}$, see Equation~\ref{eq:montecarlo_approx_ddfa}.} The neural network parameters $\mu$, $\mathcal{T}$, and $\lambda$ are initialized from the DFA as:
\begin{equation}
\begin{aligned}
    \mu[j] &= 
    \begin{cases}
        1 & \text{if } q_j = q_0, \\
        0 & \text{otherwise},
    \end{cases} \\
    \mathcal{T}[s, q_i, q_j] &=
    \begin{cases}
        1 & \text{if } \delta(q_i, s) = q_j, \\
        0 & \text{otherwise},
    \end{cases} \\
    \lambda[j] &= 
    \begin{cases}
        1 & \text{if } q_j \in F, \\
        0 & \text{otherwise}.
    \end{cases}
\end{aligned}
\end{equation}
\textcolor{black}{
\paragraph{Running Example: DFA Tensorization}
The preprocessed DFA $A_\phi'$ shown in Figure~\ref{fig:running_example} is converted into the three-dimensional tensor $\mathcal{T}$ encoding the transition function, and into the two vectors $\mu$ and $\lambda$, which represent the initial state and the accepting states, respectively, as illustrated in the same figure. Note that a 1 in position $(i,j)$ of $\mathcal{T}[x]$ indicates a transition $q_i \xrightarrow{x} q_j$ in the preprocessed automaton.
}

\subsection{Differentiable Sampling}
For the computation of the global logic loss, our goal is to generate complete suffixes and evaluate their compliance with the knowledge constraint during training.
To this end, note that the next-activity predictor is trained on \emph{perfectly one-hot} (i.e., symbolic) input sequences $\boldsymbol{a}_{\leq t}$ and produces continuous probability vectors $\tilde{y}^{(t)}$ as output, which can differ significantly from one-hot vectors. As a result, we cannot directly feed these probability vectors back into the network as inputs in subsequent steps, as doing so may lead to unpredictable behavior. Instead, we need to \emph{sample} from these distributions to recover one-hot-like inputs. At the same time, this sampling must remain differentiable to enable backpropagation.

The computation of the global logic loss proceeds as follows:
\begin{enumerate}
    \item Given a prefix $\boldsymbol{p}_t$, generate $N$ suffixes $\boldsymbol{\tilde{s}}_t^{(i)}$ that are simultaneously:
        (i) highly probable under the network’s distribution;
        (ii) differentiable; and
        (iii) \emph{nearly symbolic}, i.e., as close as possible to one-hot vectors;
    \item Evaluate whether the generated complete traces $\boldsymbol{p}_t + \boldsymbol{\tilde{s}}_t^{(i)}$ satisfy the \ltlf formula $\phi$ using DeepDFA, which supports evaluation over both categorically grounded and probabilistically grounded traces;\footnote{Note that the prefix $\boldsymbol{p}_t$ is categorically grounded, whereas the suffix $\boldsymbol{\tilde{s}}_t$ is probabilistically grounded.}
    \item Maximize the estimated satisfaction probability $\hat{P}_{\theta \vDash \phi}$, computed as the empirical mean over the sampled traces.
\end{enumerate}

\begin{figure}[t!]
    \centering
\includegraphics[width=1\linewidth]{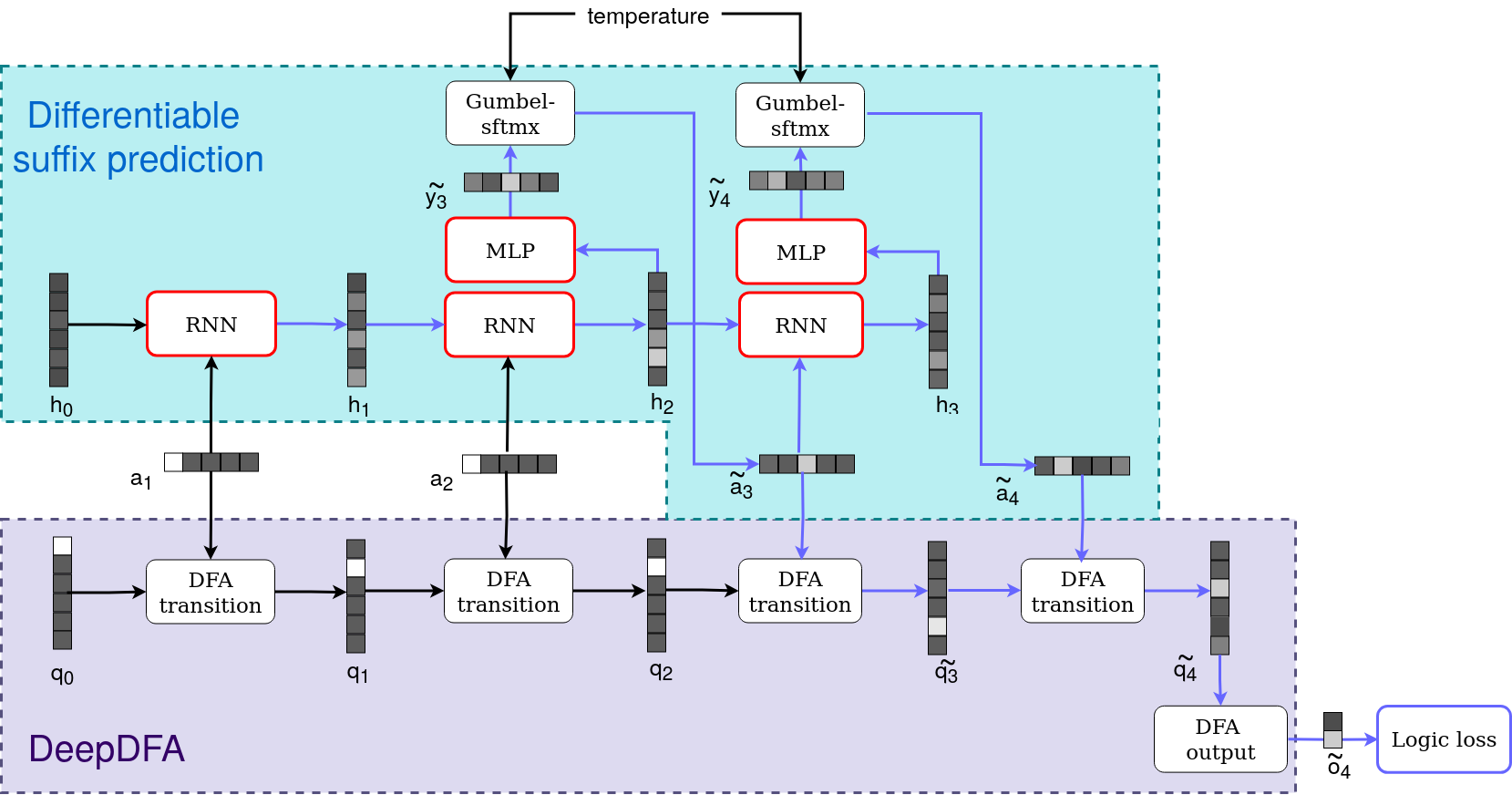}
    \caption{Global logic loss computation using a differentiable procedure for both suffix generation and formula evaluation. Violet arrows indicate the connections through which the loss is back-propagated, while components highlighted with a red border denote the modules whose parameters are updated by the logic loss. \textcolor{black}{The figure is the same as in \cite{UmiliPMAI24}, with adapted notation.}}
    \label{fig:logic_loss}
\end{figure}

Thanks to the differentiability of both the knowledge evaluator and the sampling process, the resulting loss can be back-propagated through the generated suffixes and used to update the parameters $\theta$ of the next-activity predictor, as illustrated in Figure~\ref{fig:logic_loss}.
To sample the next activity $\tilde{a}_t$ from the probability distribution $\tilde{y}_t$ produced by the network (while ensuring differentiability), we employ the Gumbel-Softmax reparameterization trick~\cite{gumb_sftmx_1, gumb_sftmx_2}:
\begin{equation} \label{eq:diff_sampling}
    \tilde{a}_t = \text{softmax}\left(\frac{\log(\tilde{y}_t) + G}{\tau}\right),
\end{equation}
where $G$ is a random vector sampled from the Gumbel distribution, and $\tau$ is a temperature parameter that controls the sharpness of the output distribution. As $\tau \rightarrow 0$, the output approaches a discrete one-hot vector, while for $\tau = 1$, it remains close to the original continuous probabilities in $\tilde{y}_t$. Since the next activity is only \emph{probabilistically} grounded, we denote it as $\tilde{a}_t$.

\textcolor{black}
{\paragraph{Running Example: Local and Global Loss Computation}
We now have the full theoretical framework for computing the two loss functions for our running example characterized by the automaton $A_\phi'$ of Figure~\ref{fig:running_example}.
}

\textcolor{black}{The local loss is computed under the \emph{teacher forcing} training procedure, meaning that at each time step, the model is conditioned on the ground-truth prefix of the sequence and penalized for assigning probability mass to labels that would cause a permanent violation. To compute this penalty, the prefix is simulated step by step through the DFA in order to identify the labels that would lead to a rejecting state and to sum the probability mass assigned to them. Consider the following sequence, which satisfies the automaton $A_\phi'$:
\[
\boldsymbol{a} = (c,a,c,b,\texttt{EOT}).
\]
At each time step, the current prefix is simulated through the automaton, and the model outputs a probability distribution over the alphabet:
\begin{align*}
\boldsymbol{a}_{<1} &= () 
& f_\theta(\boldsymbol{a}_{<1}) &= [p(a), p(b), p(c), p(\texttt{EOT})] = [0.35,\,0.15,\,0.4,\,0.1],\\
\boldsymbol{a}_{<2} &= (c) 
& f_\theta(\boldsymbol{a}_{<2}) &= [0.55,\,0.25,\,0.15,\,0.05],\\
\boldsymbol{a}_{<3} &= (c,a) 
& f_\theta(\boldsymbol{a}_{<3}) &= [0.1,\,0.65,\,0.1,\,0.15],\\
\boldsymbol{a}_{<4} &= (c,a,c) 
& f_\theta(\boldsymbol{a}_{<4}) &= [0.1,\,0.15,\,0.55,\,0.2],\\
\boldsymbol{a}_{<5} &= (c,a,c,b) 
& f_\theta(\boldsymbol{a}_{<5}) &= [0.05,\,0.1,\,0.15,\,0.7].
\end{align*}
\indent The automaton $A_\phi'$ encodes the Response Declare constraint that has the liveness property. Recall that such a constraint can be permanently violated only if the sequence terminates before the constraint is satisfied. In this example, once the symbol $a$ has been observed, the constraint requires that $b$ must eventually follow before the sequence ends. Therefore, at prefixes where $a$ has occurred but $b$ has not yet been seen (steps 3 and 4), predicting \texttt{EOT} would cause a permanent violation. In these cases, the invalid probability mass corresponds solely to the probability assigned to \texttt{EOT}. The violating mass at each time step is computed using Equation~\ref{eq:irreversible_violation_prob} and is therefore:
\begin{align*}
P(\boldsymbol{a}_1 \nvDash^{\times} \phi \mid \boldsymbol{a}_{<1}) &= 0,\\
P(\boldsymbol{a}_2 \nvDash^{\times} \phi \mid \boldsymbol{a}_{<2}) &= 0,\\
P(\boldsymbol{a}_3 \nvDash^{\times} \phi \mid \boldsymbol{a}_{<3}) &= p(\texttt{EOT}) = 0.15,\\
P(\boldsymbol{a}_4 \nvDash^{\times} \phi \mid \boldsymbol{a}_{<4}) &= p(\texttt{EOT}) = 0.2,\\
P(\boldsymbol{a}_5 \nvDash^{\times} \phi \mid \boldsymbol{a}_{<5}) &= 0.
\end{align*}
\indent After computing the invalid mass at each step, the local loss is obtained by applying Equation~\ref{eq:lll}, which averages the negative logarithm of the valid mass across all time steps:
\begin{align*}
L_\phi^{\text{loc}}
&= \frac{1}{5}\Bigl[0 + 0 -\log(1-0.15) - \log(1-0.2) + 0 \Bigr] \\[4pt]
&= \frac{-\log(0.85)-\log(0.8)}{5} \approx 0.0771.
\end{align*}
}

\textcolor{black}{
Regarding the global loss, we recall that this loss is computed as follows: (i) we generate \(N\) traces of length \(T\), and (ii) we compute over them the probability that the autoregressive model satisfies the specification encoded by the automaton, denoted as $\hat{P}_{\theta \vDash \phi}$ and defined in Equation~\ref{eq:montecarlo_approx_ddfa}.
Consider the case where we simulate three traces of length four (\(N=3, T=4\)):
\[
\boldsymbol{a}^{(1)} = (a,b,b,b), \quad
\boldsymbol{a}^{(2)} = (a,c,b,\texttt{EOT}), \quad
\boldsymbol{a}^{(3)} = (a,a,\texttt{EOT},b).
\]
\indent Among these, only \(\boldsymbol{a}^{(2)}\) satisfies the automaton \(A_\phi'\). Indeed, it is the only trace that satisfies the specification \textit{and then} properly terminates. Trace \(\boldsymbol{a}^{(1)}\) satisfies the specification but does not terminate, while \(\boldsymbol{a}^{(3)}\) terminates one step before \(T\) but does not satisfy the specification prior to termination. The satisfaction probability is therefore
\[
P_{\phi} = \frac{1}{3}(0 + 1 + 0) = 0.33.
\]
\indent For simplicity, this example assumes perfectly categorical sampling of symbols. In practice, however, sampling is performed via the Gumbel--Softmax in order to preserve differentiability. The Gumbel--Softmax \emph{approximates} categorical sampling; thus, the traces \(\boldsymbol{a}^{(i)}\) may in practice be sampled as follows (in bold the highest probability value): \newline \begin{minipage}{0.3\textwidth} \small \begin{align*} \tilde{\boldsymbol{a}}^{(1)} =& ([\textbf{0.99},0,0.01,0],\\ &[0.05,\textbf{0.82},0.05,0.08],\\ &[0.05,\textbf{0.9},0.05,0],\\ &[0,\textbf{0.75},0.05,0.2]) \end{align*} \end{minipage} \begin{minipage}{0.3\textwidth} \small \begin{align*} \tilde{\boldsymbol{a}}^{(2)} =& ([\textbf{0.99},0,0.01,0],\\ & [0,0.02,\textbf{0.98},0],\\ & [0,\textbf{1},0,0],\\ & [0,0.2,0,\textbf{0.8}]) \end{align*} \end{minipage} \begin{minipage}{0.3\textwidth} \small \begin{align*} \tilde{\boldsymbol{a}}^{(3)} = &([\textbf{0.99},0,0.01,0],\\ &[\textbf{0.96},0.02,0.02,0],\\ &[0.1,0.2,0.04,\textbf{0.66}],\\ &[0.05,\textbf{0.7}, 0.2,0.05]), \end{align*} \end{minipage}\\ \newline
where each time step is represented by a probability vector over the alphabet, whose largest component corresponds to symbol \(a\) at the first step, \(c\) at the second, and so on. In this case, the probability that this probabilistic grounding of the trace satisfies the specification is no longer exactly one, but remains close to it. DeepDFA naturally handles such probabilistic groundings.\\
\indent We now compute the satisfaction probability $P_{\mathrm{DDFA}}(\tilde{\boldsymbol{a}}^{(i)} \vDash \phi)$ by applying Equation~\ref{eq:deepDFA} using the matrices shown in Figure~\ref{fig:running_example}. 
$P_{\mathrm{DDFA}}(\tilde{\boldsymbol{a}}^{(1)} \vDash \phi)$ is computed as:
\begin{align*} 
\tilde{q}_1 &= [0.01, 0.99, 0, 0],\\
\tilde{q}_2 &= [0.82, 0.1, 0, 0.08],\\
\tilde{q}_3 &= [0.87, 0.05, 0, 0.08],\\
\tilde{q}_4 &= [0.73, 0, 0.17, 0.09],\\
P_{\mathrm{DDFA}}(\tilde{\boldsymbol{a}}^{(1)} \vDash \phi) = \tilde{o}_4 &= 0.17.
\end{align*}
$P_{\mathrm{DDFA}}(\tilde{\boldsymbol{a}}^{(2)} \vDash \phi)$ is computed as:
\begin{align*} 
\tilde{q}_1 &= [0.01, 0.99, 0, 0],\\
\tilde{q}_2 &= [0.03, 0.97, 0, 0],\\
\tilde{q}_3 &= [1, 0, 0, 0],\\
\tilde{q}_4 &= [0.2, 0, 0.8, 0],\\
P_{\mathrm{DDFA}}(\tilde{\boldsymbol{a}}^{(2)} \vDash \phi) = \tilde{o}_4 &= 0.8.
\end{align*}
$P_{\mathrm{DDFA}}(\tilde{\boldsymbol{a}}^{(3)} \vDash \phi)$ is computed as:
\begin{align*} 
\tilde{q}_1 &= [0.01, 0.99, 0, 0],\\
\tilde{q}_2 &= [0.02, 0.98, 0, 0],\\
\tilde{q}_3 &= [0.2, 0.14, 0.01, 0.65],\\
\tilde{q}_4 &= [0.28, 0.05, 0.02, 0.66],\\
P_{\mathrm{DDFA}}(\tilde{\boldsymbol{a}}^{(3)} \vDash \phi) = \tilde{o}_4 &= 0.02.
\end{align*}
The final numeric value of the GLL is therefore
\[
L_\phi^{\text{glob}} = -\log\left(\frac{0.17 + 0.8 + 0.02}{3}\right) = 1.11.
\]
}

\subsection{Applications beyond PPM and Limitations} \label{sec:limitations}
\textcolor{black}{
In this paper, we focus on integrating logical and autoregressive losses in the context of PPM. However, our approach is general and can be applied to other domains, which we do not address here and leave for future work.
Indeed, suffix prediction in PPM is simply a specific instance of discrete sequence generation. Many applications fall under this same paradigm, including language generation in NLP~\cite{transformers} and neural machine translation~\cite{few_shot_learner}. Given the remarkable success of Transformers across a wide range of tasks, a current trend is to employ discrete autoregression even in originally continuous domains, after discretization or ``tokenization''~\cite{trajectory_transformers}. Across all these applications, our framework provides a principled way to integrate prior knowledge -- expressed as temporal formulas or automata -- into sequential data modeling.
}

\textcolor{black}{
That said, PPM tasks typically involve medium-scale problem sizes (especially when compared to LLMs), in terms of both vocabulary size and trace length. The imposed specifications are also of reasonable complexity~\footnote{In our datasets, the number of activities and the trace length range from 10 to 50 and from 10 to 100, respectively. The number of states of the automata varies between 6 and 13.}. Applying our approach to substantially larger domains or more complex specifications may introduce the following challenges, which we summarize below together with possible mitigation strategies:}
\begin{enumerate}
    \item 
    \textcolor{black}{
    \textbf{Large vocabularies.}  
    As discussed in Section~\ref{sec:knowledge_preprocessing}, in the current version of the framework the automaton is preprocessed so as to explicitly include all symbols in the vocabulary $\mathcal{A}$. As a result, for very large vocabularies the size of the DeepDFA model can grow substantially. However, only the symbols appearing in the knowledge alphabet $\Sigma$ are actually required to check specification satisfaction, and typically $|\Sigma| \ll |\mathcal{A}|$. This issue can therefore be addressed by mapping all symbols irrelevant to the specifications into a single special token, thereby minimally increasing the size of the automaton model.}
    \item 
    \textcolor{black}{
    \textbf{Very long sequences.}  
    Recall that, to compute the global logic loss, suffix prediction must be carried out until the end of the sequence (or up to a maximum number of steps) during training. Consequently, applying the framework to very long sequences may significantly slow down the computation of the global loss. This limitation can be mitigated in different ways depending on the application domain. In natural language processing, for instance, one may segment sentences into sub-sentences, each expressing a distinct \textit{concept}~\cite{large_concept_models}, and formulate the knowledge over a concept-level vocabulary rather than over individual tokens. Similar forms of \textit{hierarchical} aggregation may also be meaningful in reinforcement learning settings~\cite{trajectory_transformers, HRL}. Ultimately, the most appropriate strategy depends on the specific domain.
    }

    \item 
    \textcolor{black}{
    \textbf{Large automata/complex specifications.}  
    Depending on the complexity of the temporal specification we want to enforce on the dataset, we may encounter: (i) long conversion times to build the automaton and/or (ii) very large automata. Indeed, given an \(\ltlf\) formula \(\phi\), theoretical results show that the size of the equivalent automaton \(A_\phi\) is double-exponential in \(\phi\) in the worst case~\cite{LTLf}. However, scalable techniques are available for computing \(A_\phi\) from \(\phi\)~\cite{LTL2DFA1, LTL2DFA2, LTL2DFA3}, and in practice the resulting automaton is often quite small and can be constructed in a reasonable amount of time. This is especially true for Declare formulas, which have been shown to generate DFAs that are polynomial in the size of the original formula~\cite{declare2011}. Moreover, note that automaton construction in our framework is performed \emph{offline} and only once, so its computation time does not significantly affect overall performance. For very complex knowledge bases, conversion times and automaton size can be further reduced by representing the specification as a conjunction of simpler formulas (e.g., \(\phi = \phi_1 \wedge \phi_2 \wedge \phi_3\)), converting each subformula into its own automaton (\(A_{\phi_1}, A_{\phi_2}, A_{\phi_3}\)), and monitoring the states of all automata in parallel. The knowledge is considered satisfied if all automata reach a final state at the end of the trace.
    }
\end{enumerate}
\textcolor{black}{
These considerations provide useful guidelines for extending the framework to substantially larger domains. They are not required in our current experiments, as the proposed approach is well suited to PPM-sized problems, and we leave their systematic exploration to future work.}

\section{Experiments}
\label{sec:exp}
This section describes the experimental setup, including the datasets, evaluation metrics, and comparative approaches, followed by a detailed analysis of the results obtained.
The experiments are reproducible using our implementations of the GLL and LLL at \href{https://github.com/axelmezini/nesy-suffix-prediction-dfa}{https://github.com/axelmezini/nesy-suffix-prediction-dfa}.

\subsection{Experimental Setup}

Our methods have been evaluated using three real-world datasets: the Sepsis event log~\footnote{\url{http://doi.org/10.4121/uuid:915d2bfb-7e84-49ad-a286-dc35f063a460}, checked on June 2026}, the BPI Challenge 2013 (closed problems)~\footnote{\url{http://doi.org/10.4121/uuid:c2c3b154-ab26-4b31-a0e8-8f2350ddac11}, checked on June 2026}, and the BPI Challenge 2020 (travel permit) event log~\footnote{\url{http://doi.org/10.4121/uuid:52fb97d4-4588-43c9-9d04-3604d4613b51}, checked on June 2026}.
Given an event log, the traces are ordered by the timestamp of the first event and split into training and test sets using an 80--20 ratio. Knowledge is extracted from the test set in the form of Declare constraints using the Declare Miner~\cite{DBLP:conf/icpm/AlmanCHMN20}, each with a minimum support value of $85\%$, that is, they are satisfied in at least $85\%$ of the traces. \textcolor{black}{This value allows us to obtain a reasonable number of constraints and test traces that satisfy them.} These constraints are then translated into their corresponding \ltlf formulas and combined into an \ltlf model using conjunctions. To enable controlled experiments, traces that violate the extracted \ltlf model (i.e., that do not satisfy all the discovered constraints) are removed from both the training and test sets.
\begin{table}[ht]
\centering
\begin{tabular}{l c c c}
\toprule
 & Sepsis & BPIC 2013 & BPIC 2020 \\
\midrule
\textcolor{black}{\# Training traces} & \textcolor{black}{547} & \textcolor{black}{590} & \textcolor{black}{1665} \\
\textcolor{black}{\# Training variants} & \textcolor{black}{438} & \textcolor{black}{61} & \textcolor{black}{50} \\
\textcolor{black}{\# Testing traces} & \textcolor{black}{143} & \textcolor{black}{228} & \textcolor{black}{730} \\
\textcolor{black}{\# Testing variants} & \textcolor{black}{121} & \textcolor{black}{24} & \textcolor{black}{24} \\
\# Activities & 16 & 7 & 51 \\
\# Constraints & 39 & 7 & 56 \\
\# DFA states & 10 & 6 & 13 \\
\# Failure states & 2 & 2 & 2 \\
\textcolor{black}{\# DFA transitions} & \textcolor{black}{170} & \textcolor{black}{48} & \textcolor{black}{676} \\
\bottomrule
\end{tabular}
\caption{Key statistics of the datasets and background knowledge used in our experiments.}
\label{tab:dataset_details}
\end{table}

Table~\ref{tab:dataset_details} summarizes the key statistics of the datasets used in our experiments. The datasets vary in size and complexity, with the number of training traces ranging from 547 to 1665 and the number of distinct activity names ranging from 7 to 51. \textcolor{black}{The semantic complexity of the background knowledge can be inspected from the table by analyzing the complexity of the generated automaton. The number of DFA states and transitions gives an idea of the complexity of the conjunction of the Declare constraints. While all DFAs have a reasonable number of states, the DFA of BPIC 2020 has a high number of transitions, suggesting a high semantic complexity of the underlying Declare model. The Declare model of BPIC 2013, on the other hand, presents the lowest semantic complexity. The mined Declare models are available in the shared GitHub repository.} Correspondingly, the number of discovered Declare constraints and the size of the resulting deterministic finite automata (DFA) differ across datasets, reflecting their behavioral complexity. Despite these differences, all datasets include a small number of failure states in their DFAs, which indicate states violating the discovered constraints.

In addition to the training set containing only positive traces, four levels of noise are introduced at the event level: $10\%$, $20\%$, $30\%$, and $40\%$. Noise is applied by randomly replacing the activity label of selected events with other labels from the activity vocabulary. \textcolor{black}{Introducing noise into the training set after filtering makes it possible to train on non-compliant traces in a more controlled manner, providing a more realistic scenario, as one may encounter in production. Starting from a set of fully compliant traces and progressively injecting different levels of noise allows the assessment of how sensitive the extracted knowledge of a specific dataset is to random activity label perturbations.
\\ \indent Table~\ref{tab:dataset_acceptance} reports the resulting proportions of traces in the training sets that remain compliant with respect to the extracted constraints after noise injection.}
\begin{table}[ht]
\color{black}
\centering
\begin{tabular}{l c c c}
\toprule
Noise & Sepsis & BPIC 2013 & BPIC 2020 \\
\midrule
10\% & 0.534 & 0.800 & 0.513 \\
20\% & 0.296 & 0.597 & 0.255 \\
30\% & 0.135 & 0.451 & 0.094 \\
40\% & 0.044 & 0.349 & 0.048 \\
\bottomrule
\end{tabular}
\caption{\textcolor{black}{Compliance ratio of the training sets after noise injection.}}
\label{tab:dataset_acceptance}
\end{table}
\textcolor{black}{The BPIC 2020 and Sepsis datasets are particularly sensitive, retaining only slightly more than 50\% compliant traces at a noise level of 10\% and less than 5\% at a noise level of 40\%. In contrast, BPIC 2013 appears more robust, which is expected given its lower number of constraints. As a result, it maintains at least 35\% compliant traces even at the highest noise level. During GLL training and GLL and LLL testing, different prefix lengths are used: $\lfloor M/2 \rfloor$, $\lfloor M/2 \rfloor + 1$, and $\lfloor M/2 \rfloor + 2$, where $M$ is the median trace length of the training log, as done in~\cite{DiFrancescomarino17}.}

\paragraph{Compared Methods} 
\textcolor{black}{We base our evaluation on two classes of next-activity predictors: recurrent and attention-based architectures.
For the recurrent setting, we consider: (i) a baseline LSTM trained only with supervised loss, (ii) an LSTM trained with supervised and local logic loss, and (iii) an LSTM trained with supervised and global logic loss. All recurrent models are two-layer LSTMs with 100 hidden units per layer, trained using a batch size of 64 and the Adam optimizer. To assess architectural generality, we additionally evaluate Transformer-based models. The Transformer consists of an input projection layer followed by positional embeddings, a stack of Transformer encoder layers with multi-head self-attention (using GELU activation and pre-layer normalization), and a linear output projection to the activity vocabulary. A causal mask is applied to enforce autoregressive generation for GLL. We evaluate baseline, baseline+LLL, and baseline+GLL variants under the same training protocol as the LSTMs. We highlight the fact that the base models (simple LSTM and Transformer) can be considered an ablation study in which the logical loss in Equation~\ref{eq:loss-balance} has been removed.}

For each setting, two different sampling strategies are tested: temperature-based sampling (at testing time for both GLL and LLL and at training time for LLL) with temperature parameter $\tau = 0.7$ (see Equation~\ref{eq:diff_sampling}), and greedy decoding at testing time for both GLL and LLL (i.e., the activity that maximizes the next-symbol probability is chosen). For each distinct configuration, 15 runs are executed. We compare the proposed approaches using two evaluation metrics: the satisfaction rate with respect to the \ltlf\ model and the normalized Damerau--Levenshtein similarity with respect to the ground-truth traces. Both metrics range from 0 to 1, following the ``higher is better'' rule.

\subsection{Results}
\begin{figure}[t!]
  \centering

  % Sepsis
  \vspace{0.3em}\noindent\textbf{Sepsis}\par\vspace{0.3em}
  \begin{subfigure}[b]{0.45\textwidth}
    \includegraphics[width=\linewidth]{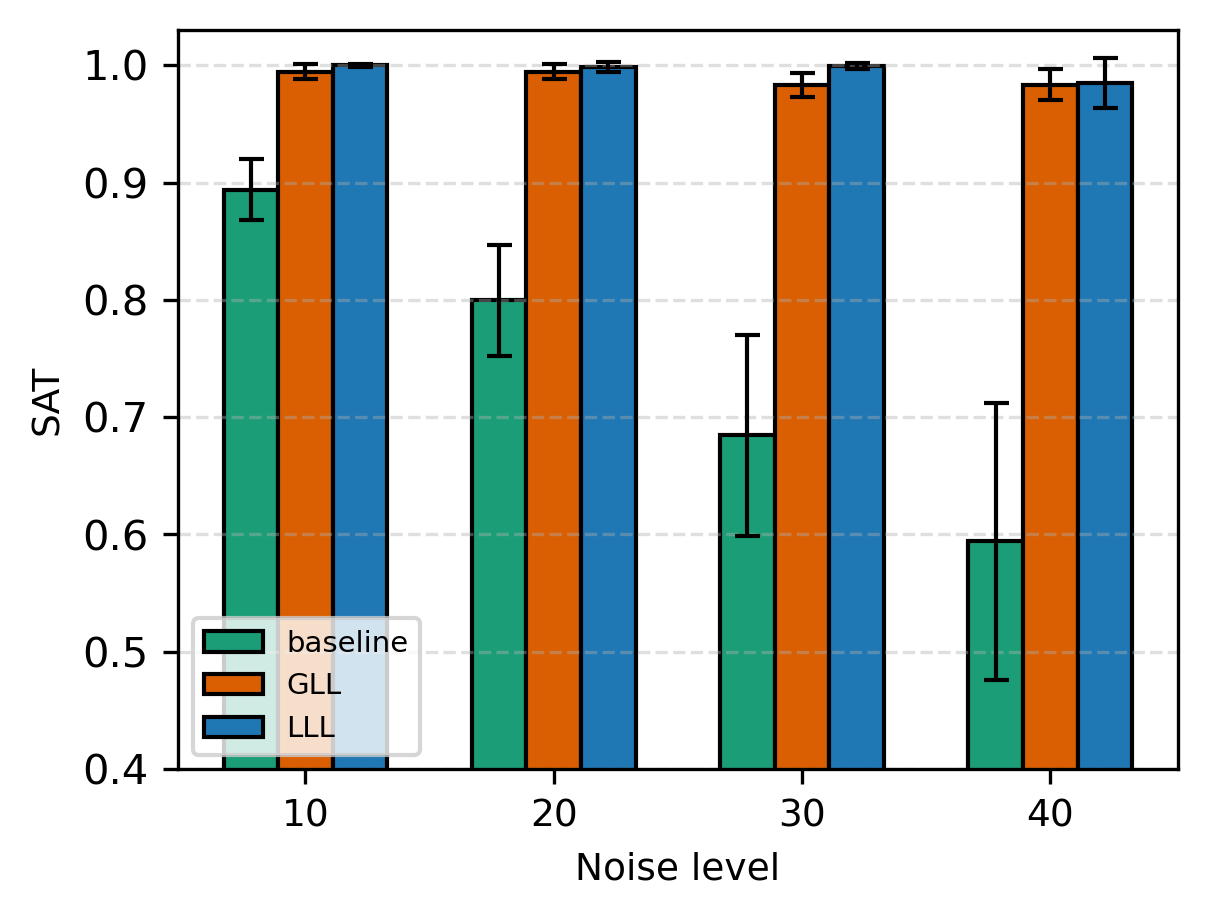}
  \end{subfigure}
  \begin{subfigure}[b]{0.45\textwidth}
    \includegraphics[width=\linewidth]{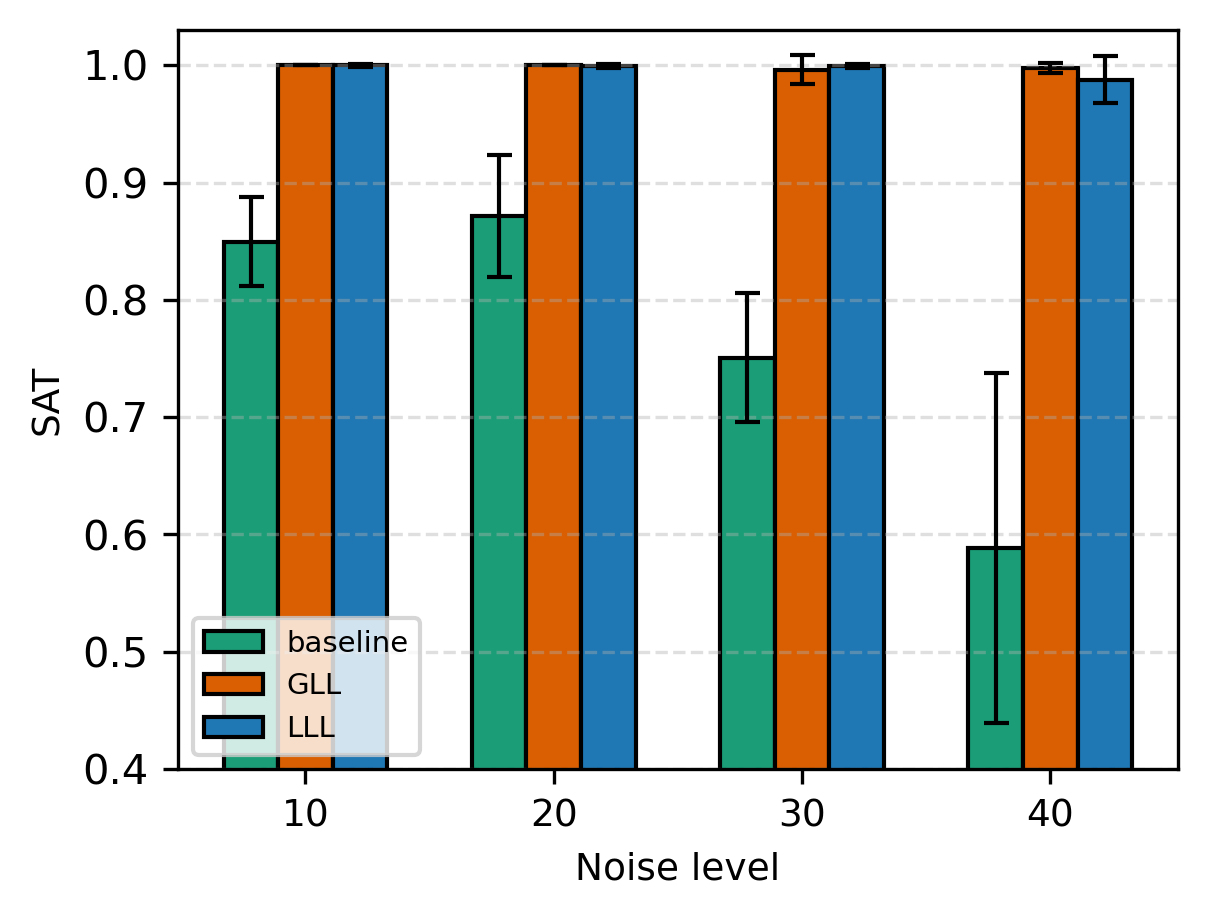}
  \end{subfigure}

  % BPIC 2013
  \vspace{0.3em}\noindent\textbf{BPIC 2013}\par\vspace{0.3em}
  \begin{subfigure}[b]{0.45\textwidth}
    \includegraphics[width=\linewidth]{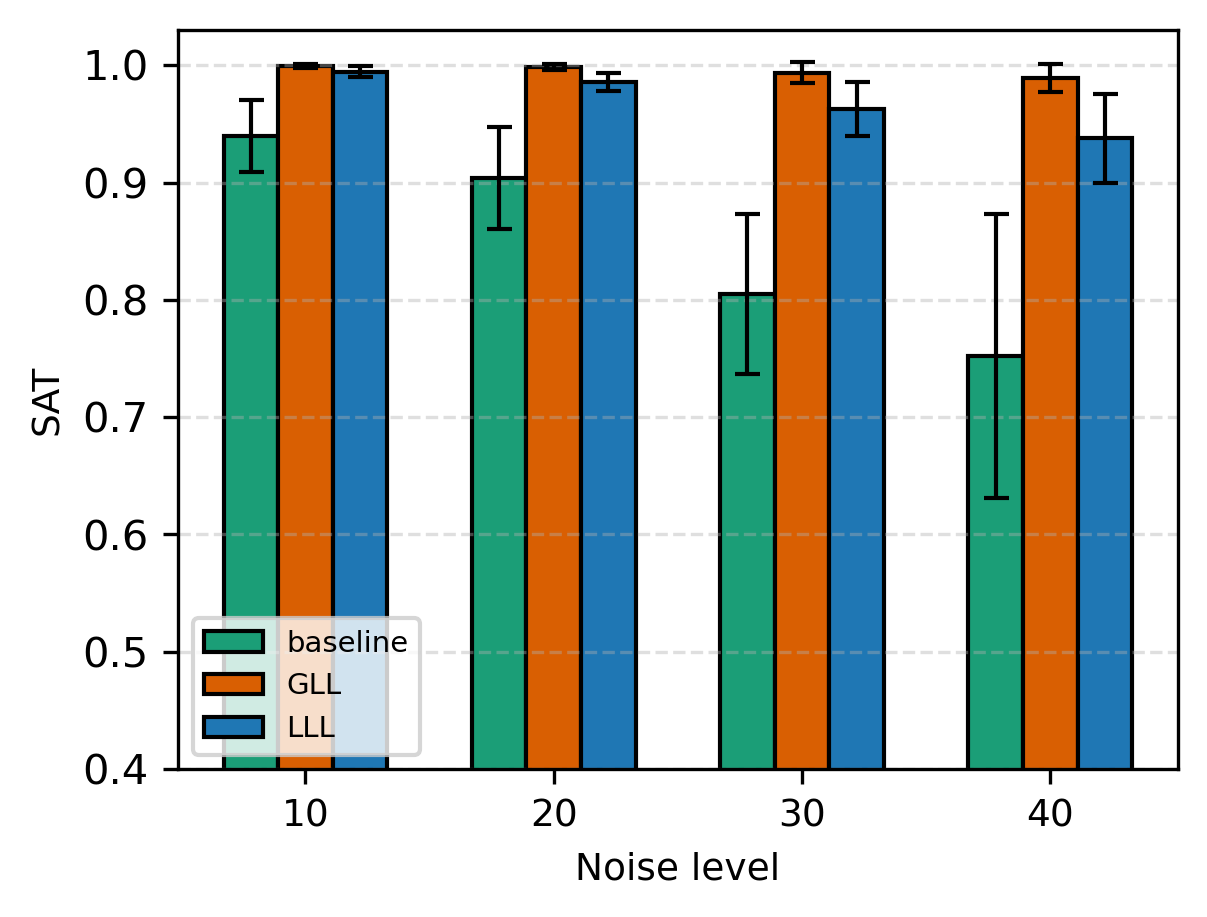}
  \end{subfigure}
  \begin{subfigure}[b]{0.45\textwidth}
    \includegraphics[width=\linewidth]{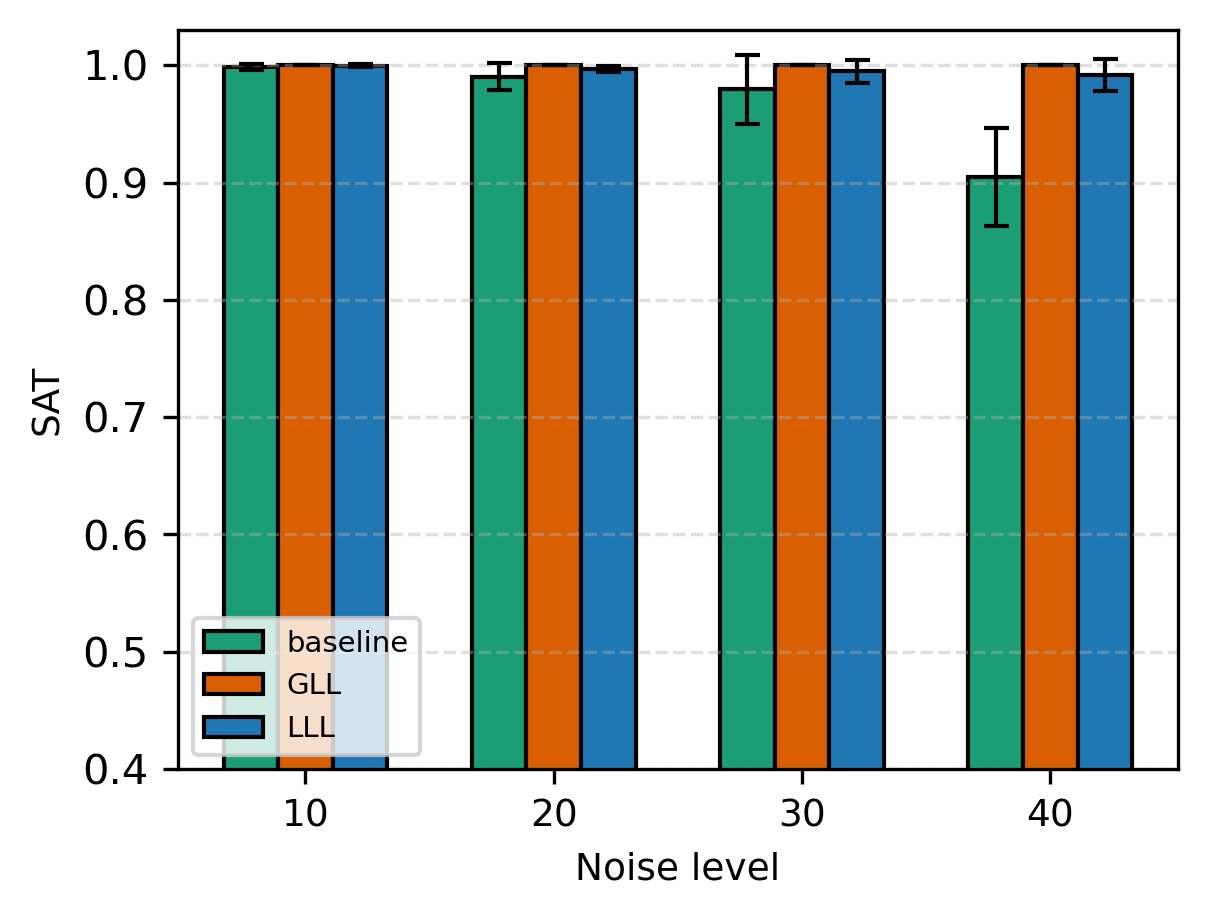}
  \end{subfigure}

  % BPIC 2020
  \vspace{0.3em}\noindent\textbf{BPIC 2020}\par\vspace{0.3em}
  \begin{subfigure}[b]{0.45\textwidth}
    \includegraphics[width=\linewidth]{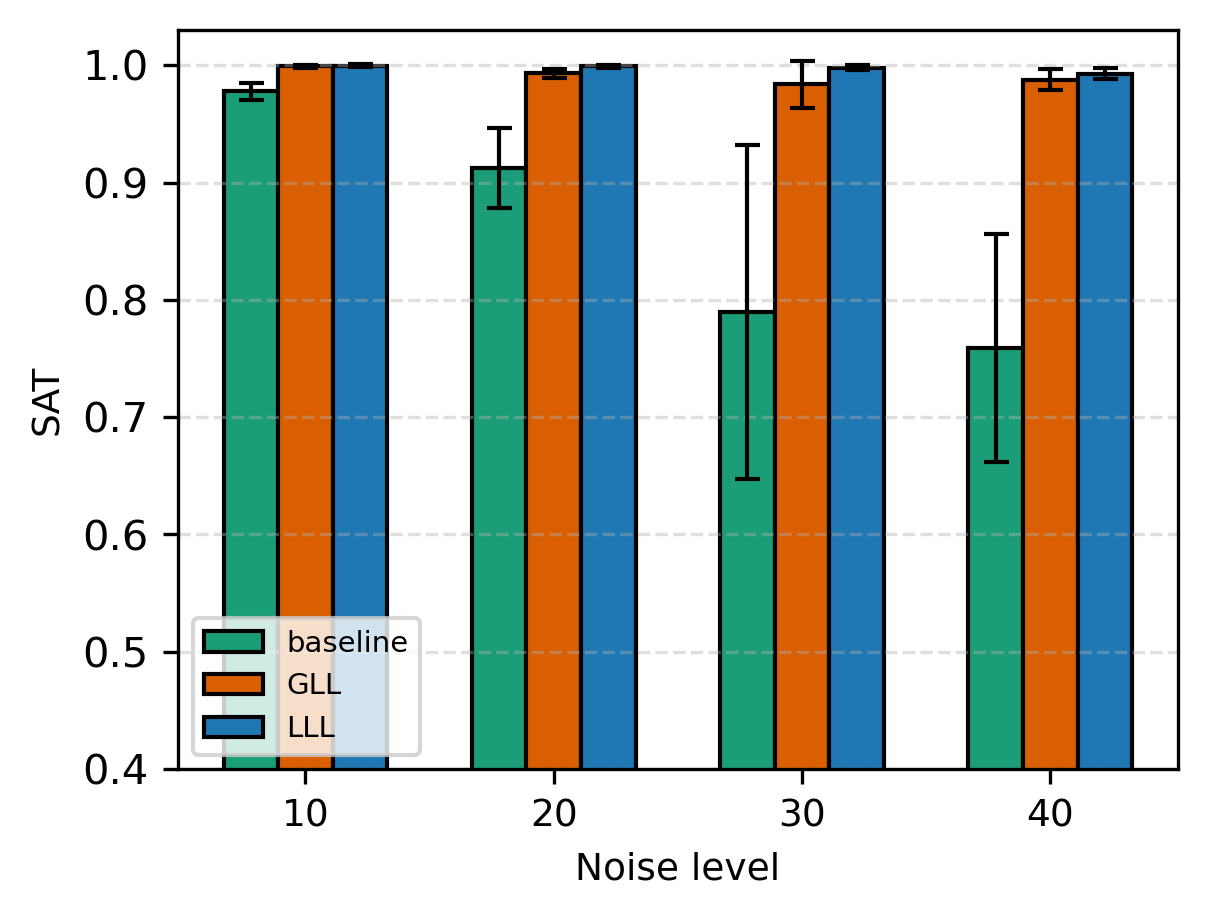}
  \end{subfigure}
  \begin{subfigure}[b]{0.45\textwidth}
    \includegraphics[width=\linewidth]{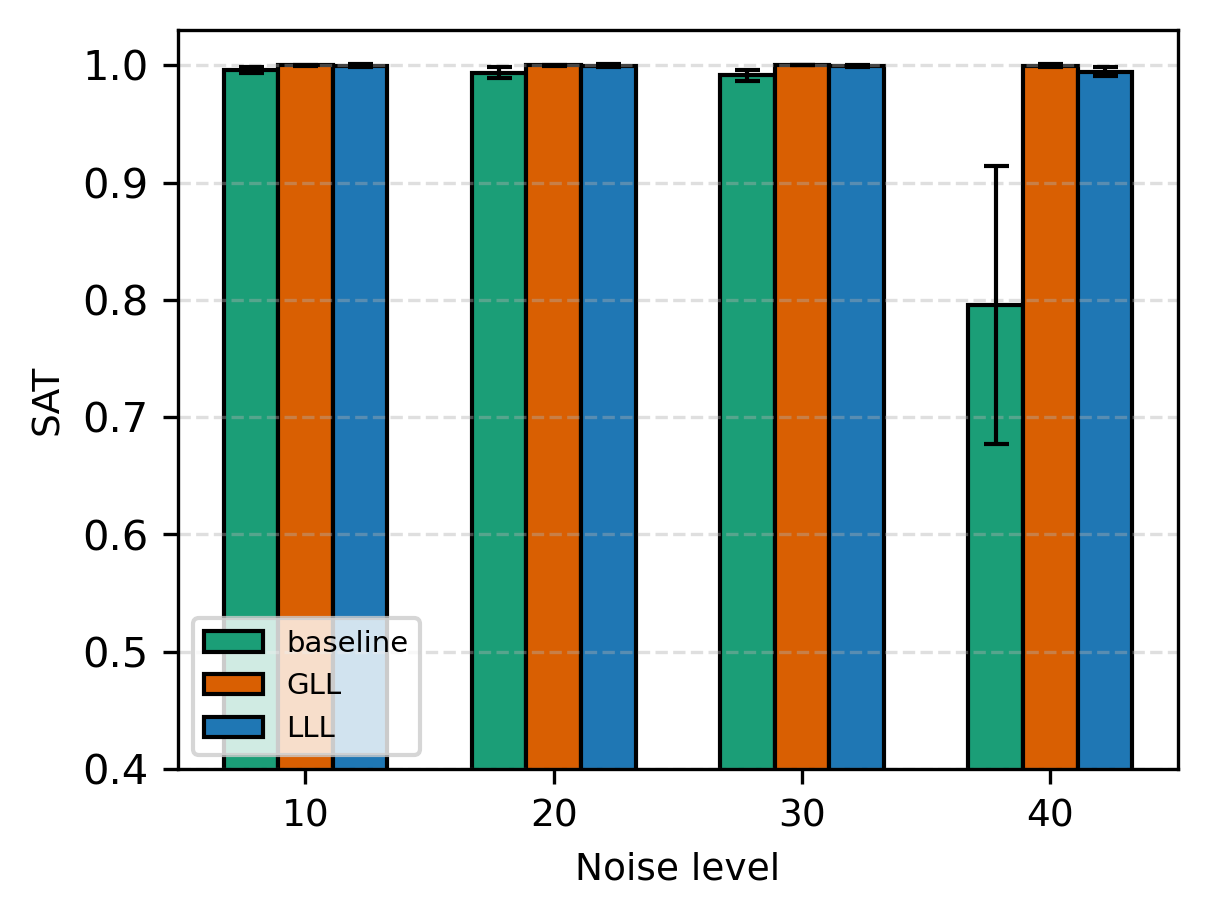}
  \end{subfigure}

  \caption{Satisfiability (SAT) of predicted traces under increasing noise levels for LSTM models. Columns correspond to the sampling strategies (temperature on the left, greedy on the right). Results are averaged over prefix lengths; error bars indicate the standard deviation over prefix lengths.}
  \label{fig:lstm_sat_comparison}
\end{figure}

\begin{figure}[!t]
  \centering

  % Sepsis
  \vspace{0.3em}\noindent\textbf{Sepsis}\par\vspace{0.3em}
  \begin{subfigure}[b]{0.45\textwidth}
    \includegraphics[width=\linewidth]{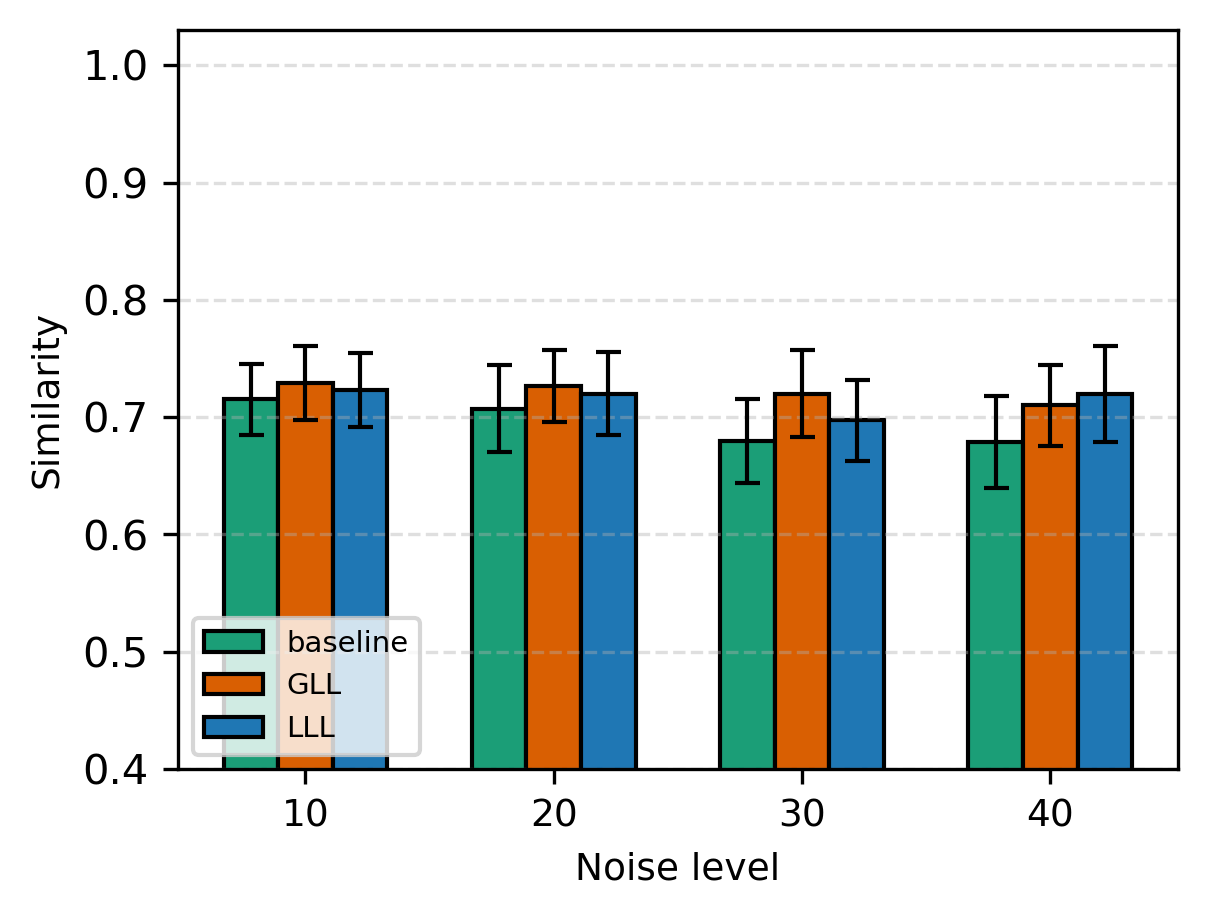}
  \end{subfigure}
  \begin{subfigure}[b]{0.45\textwidth}
    \includegraphics[width=\linewidth]{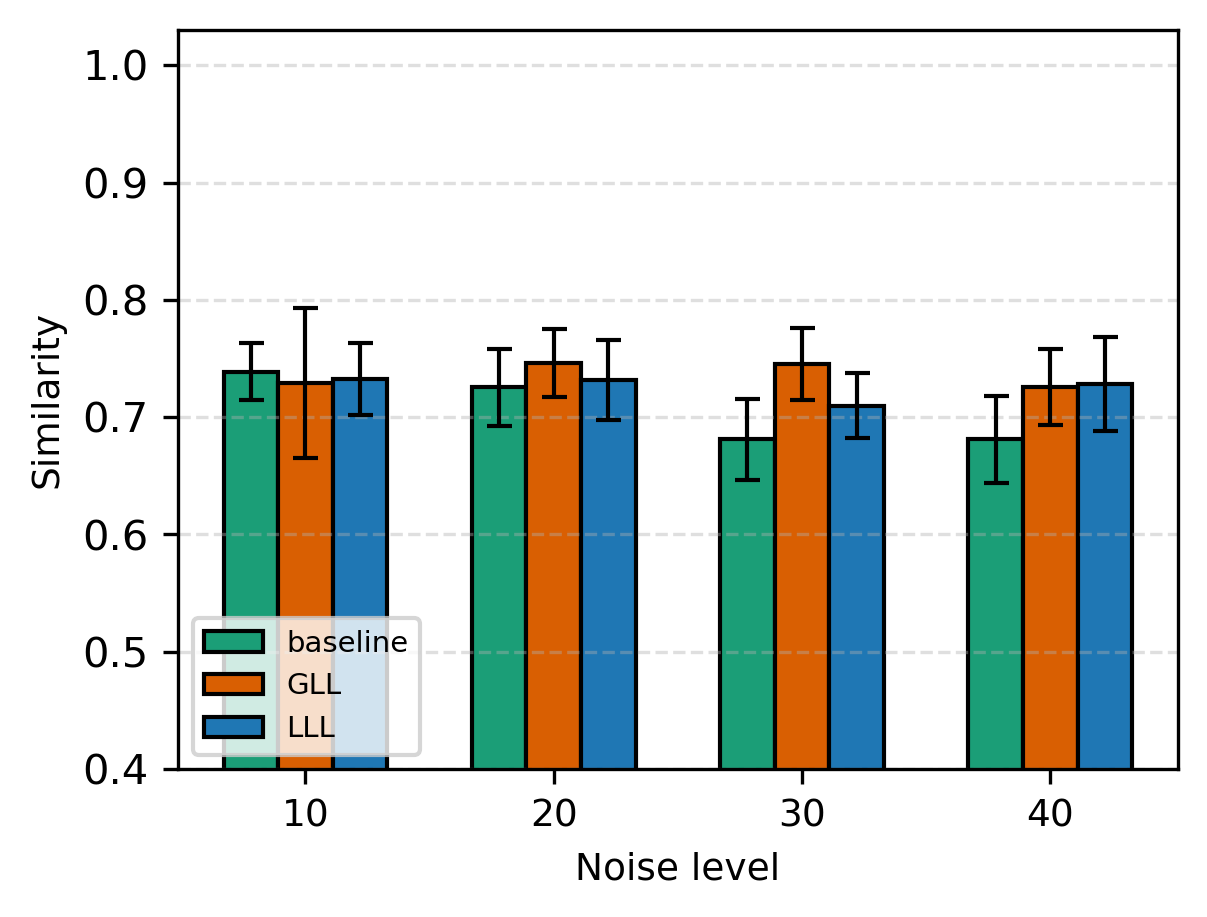}
  \end{subfigure}

  % BPIC 2013
  \vspace{0.3em}\noindent\textbf{BPIC 2013}\par\vspace{0.3em}
  \begin{subfigure}[b]{0.45\textwidth}
    \includegraphics[width=\linewidth]{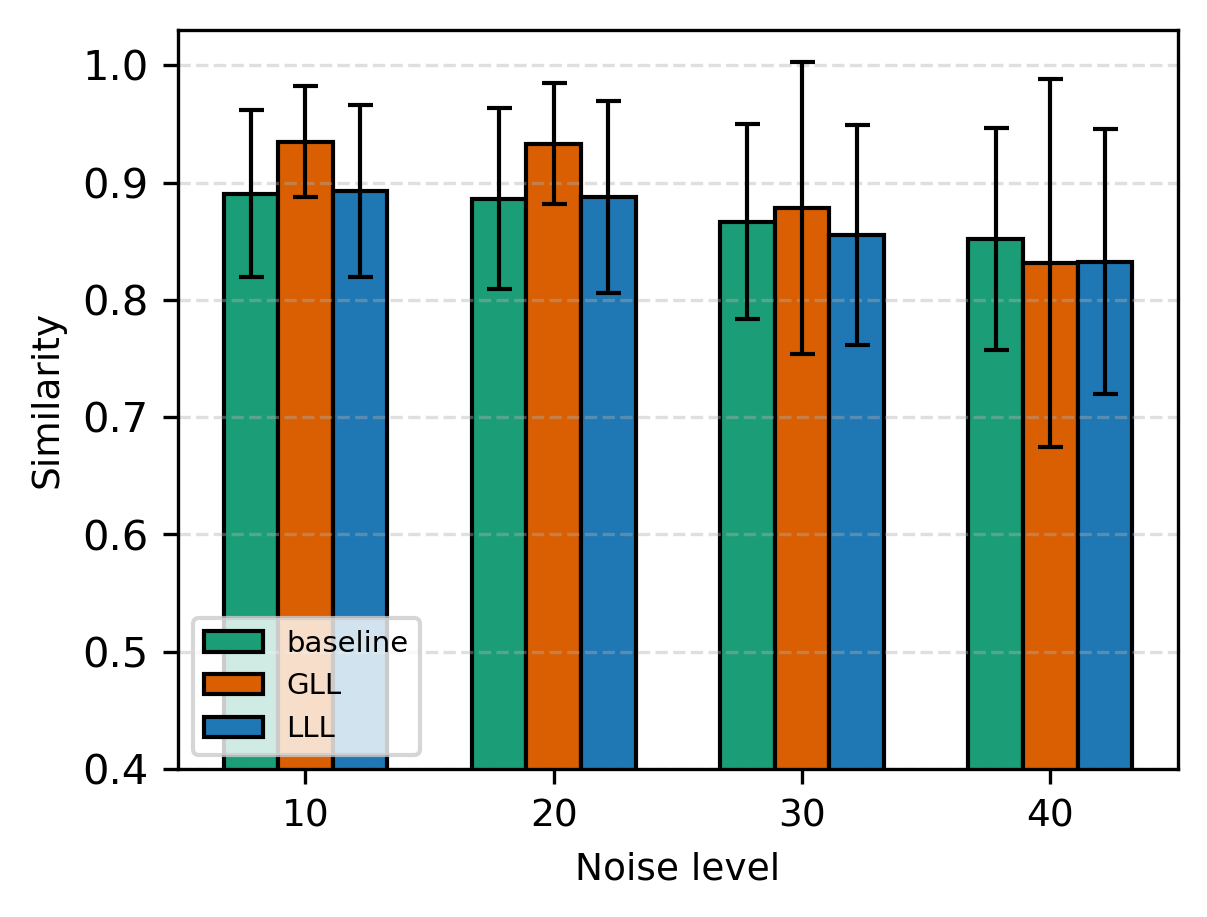}
  \end{subfigure}
  \begin{subfigure}[b]{0.45\textwidth}
    \includegraphics[width=\linewidth]{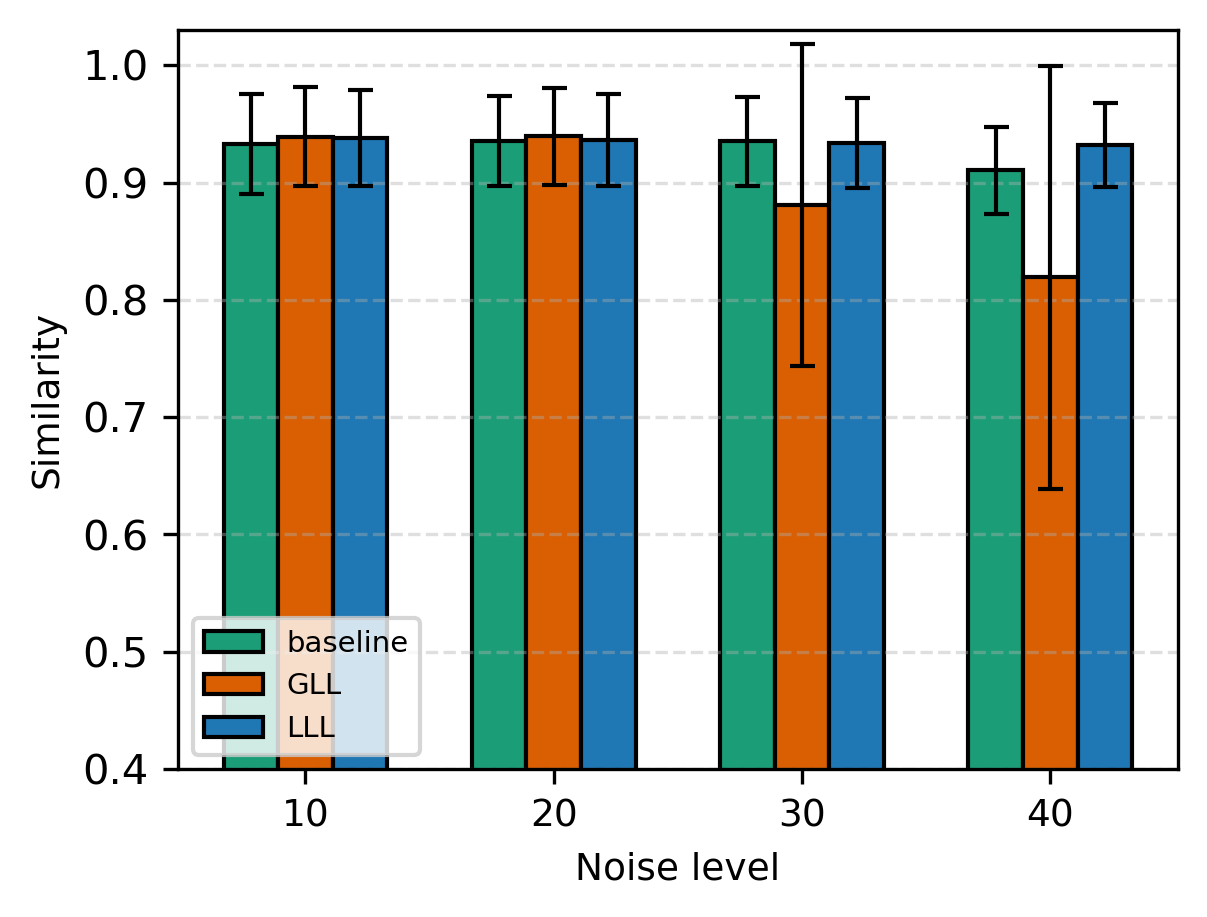}
  \end{subfigure}

  % BPIC 2020
  \vspace{0.3em}\noindent\textbf{BPIC 2020}\par\vspace{0.3em}
  \begin{subfigure}[b]{0.45\textwidth}
    \includegraphics[width=\linewidth]{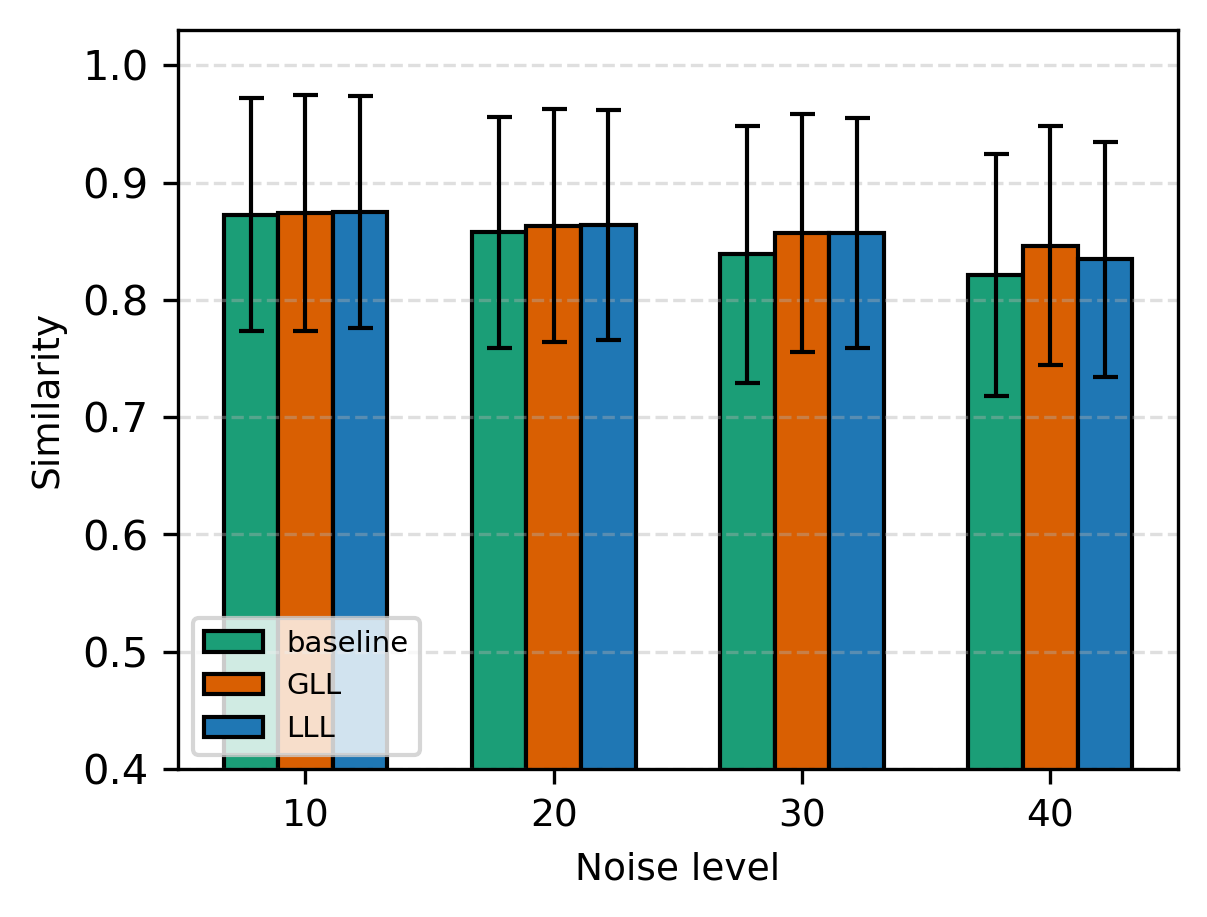}
  \end{subfigure}
  \begin{subfigure}[b]{0.45\textwidth}
    \includegraphics[width=\linewidth]{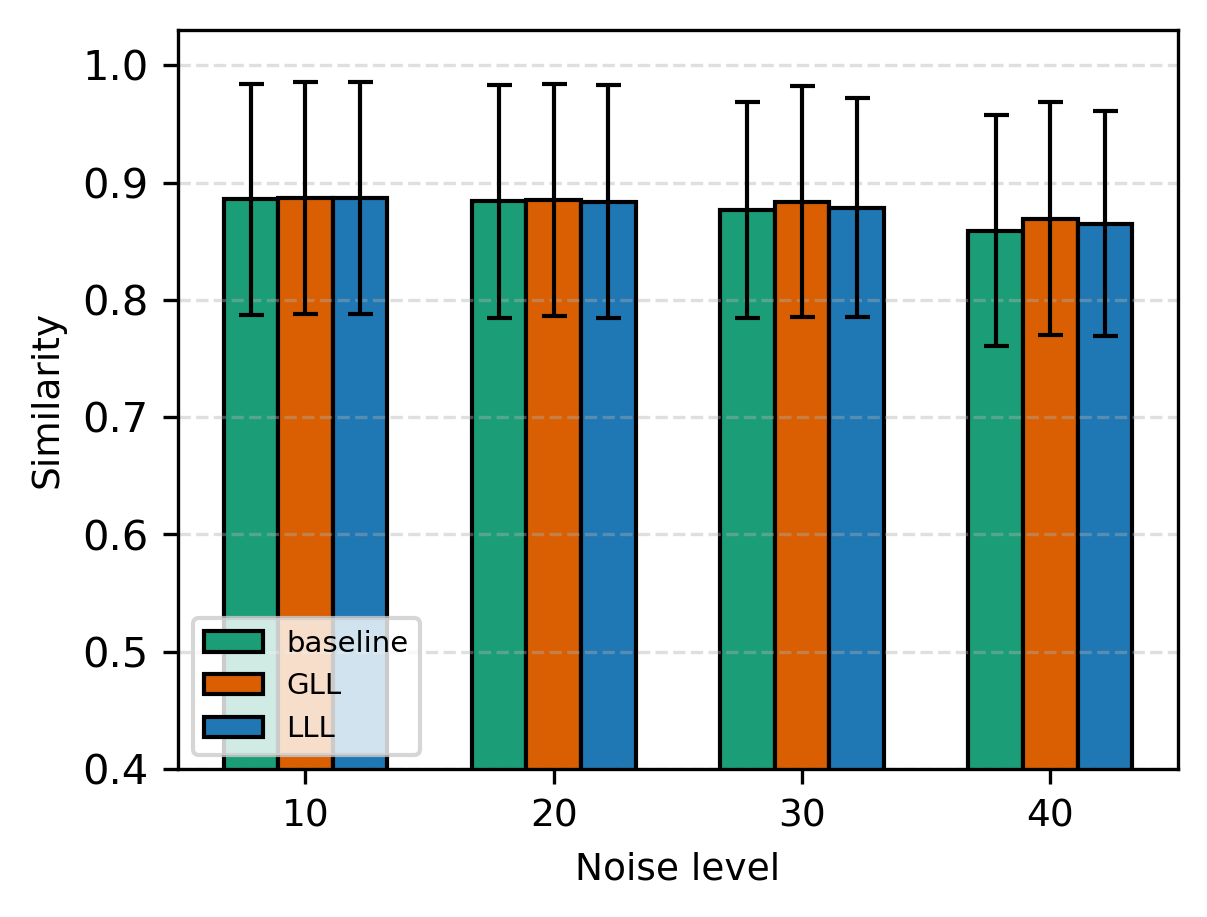}
  \end{subfigure}

  \caption{Normalized Damerau--Levenshtein similarity under increasing noise levels for LSTM models. Columns correspond to the sampling strategies (temperature on the left, greedy on the right). Results are averaged over prefix lengths; error bars indicate the standard deviation over prefix lengths.}
  \label{fig:lstm_similarity_comparison}
\end{figure}
Figures~\ref{fig:lstm_sat_comparison} and~\ref{fig:lstm_similarity_comparison} show the performance of LSTM models (with $\alpha = 0.75$ for GLL and $\alpha = 0.25$ for LLL) across all datasets and noise levels for the satisfiability and similarity metrics, respectively. The empirical results demonstrate that integrating background knowledge during training consistently increases the satisfaction rate of the predicted traces, regardless of the sampling strategy. These improvements are especially pronounced under higher noise levels, confirming the utility of incorporating logical background knowledge when predictive uncertainty increases. Notably, the satisfaction rate remains close to 100\% even at the highest noise level of $40\%$.

Importantly, this increase in satisfaction rate does not negatively affect the model’s ability to learn from the training data. The similarity to the ground-truth data remains consistent across configurations. These results suggest that the integrated knowledge helps the model distinguish compliant traces from noisy patterns in the training sets. Overall, the impact of the sampling strategy is limited, although greedy decoding often shows slightly better performance than temperature-based sampling for both metrics.

\begin{figure}[!htbp]
  \centering

  % Sepsis
  \vspace{0.3em}\noindent\textbf{Sepsis}\par\vspace{0.3em}
  \begin{subfigure}[b]{0.45\textwidth}
    \includegraphics[width=\linewidth]{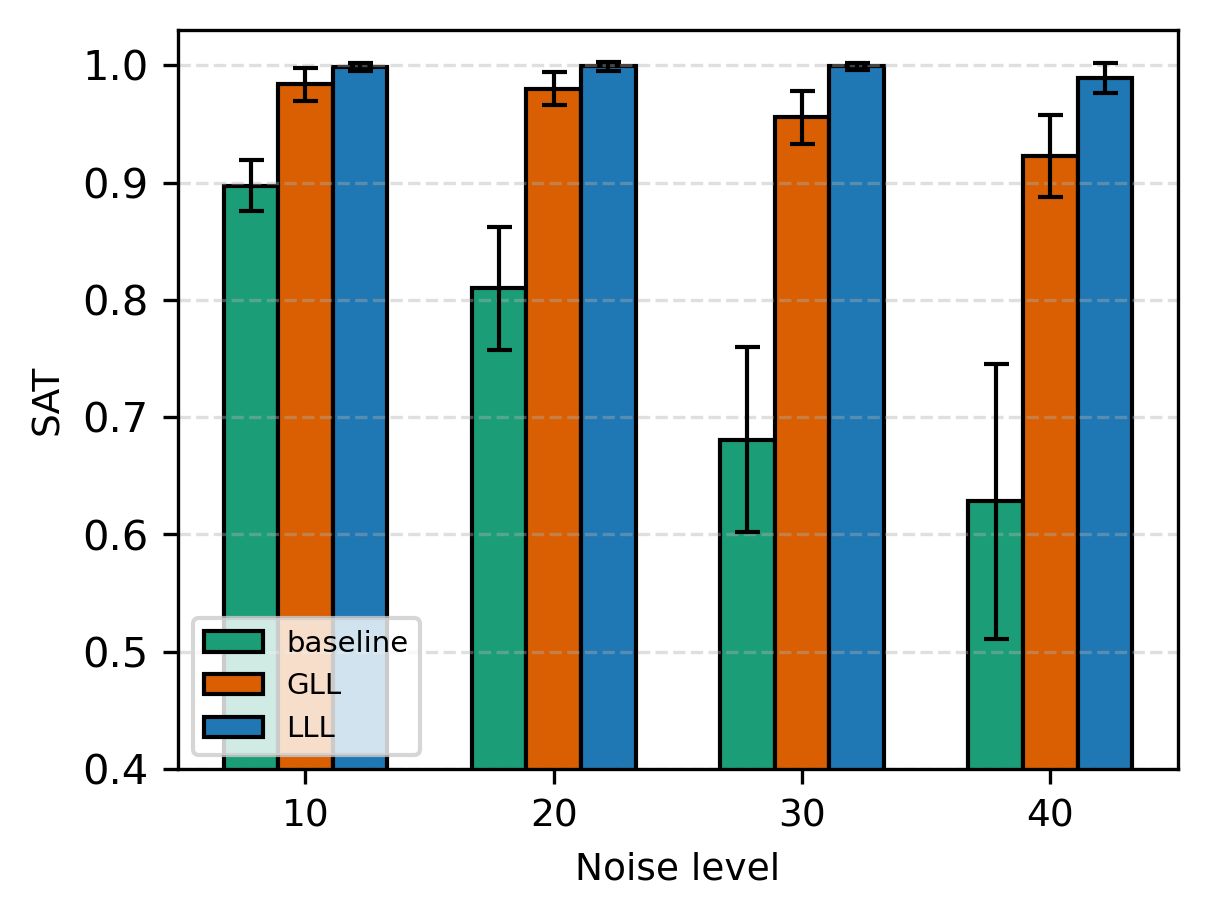}
  \end{subfigure}
  \begin{subfigure}[b]{0.45\textwidth}
    \includegraphics[width=\linewidth]{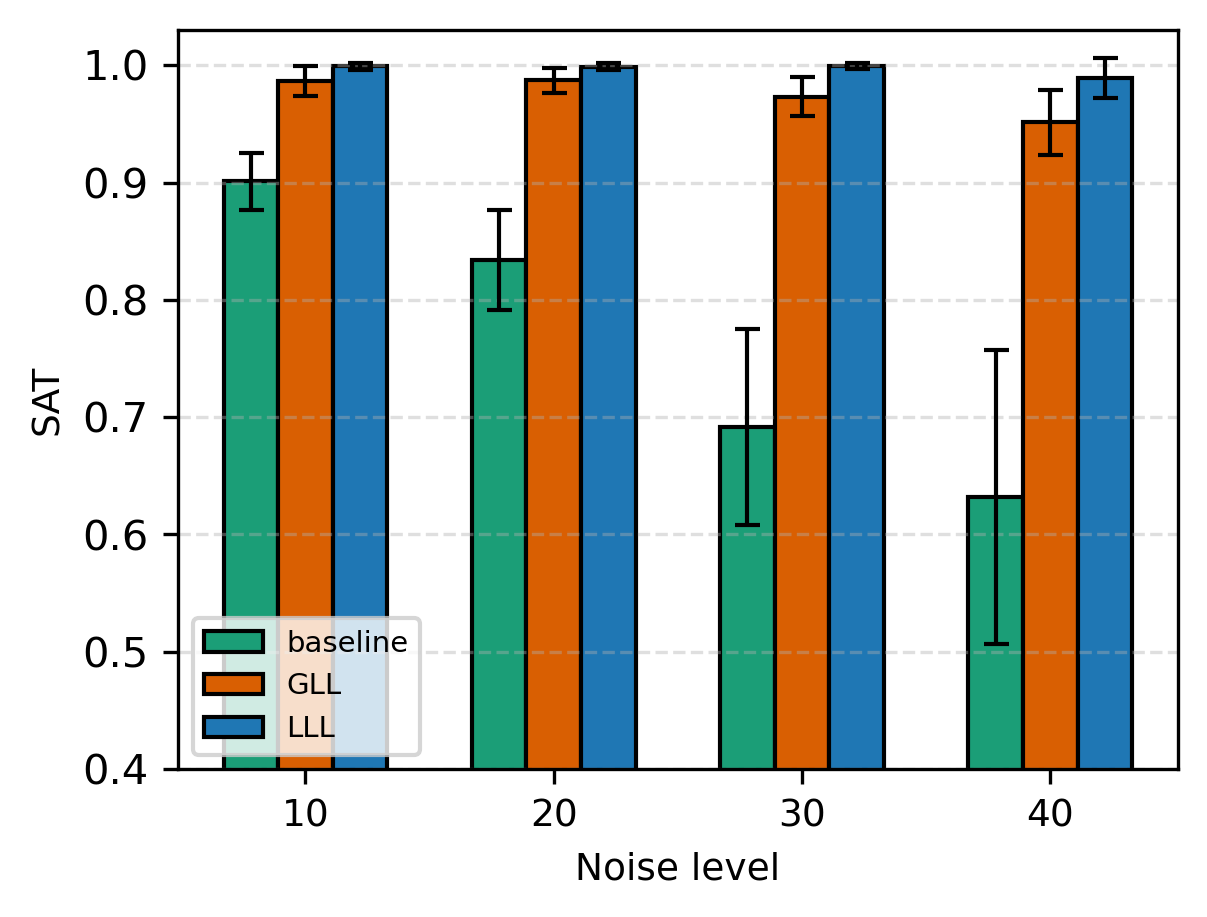}
  \end{subfigure}

  % BPIC 2013
  \vspace{0.3em}\noindent\textbf{BPIC 2013}\par\vspace{0.3em}
  \begin{subfigure}[b]{0.45\textwidth}
    \includegraphics[width=\linewidth]{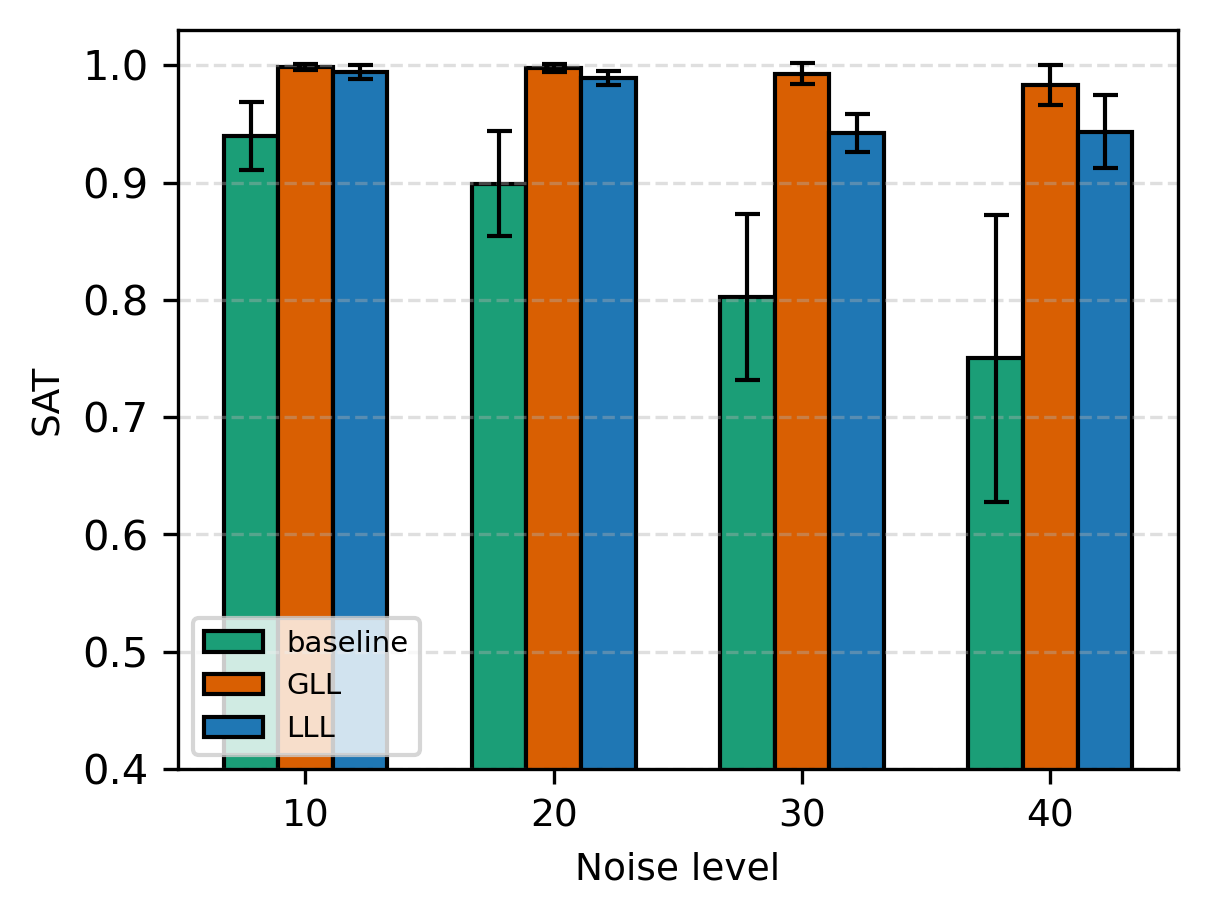}
  \end{subfigure}
  \begin{subfigure}[b]{0.45\textwidth}
    \includegraphics[width=\linewidth]{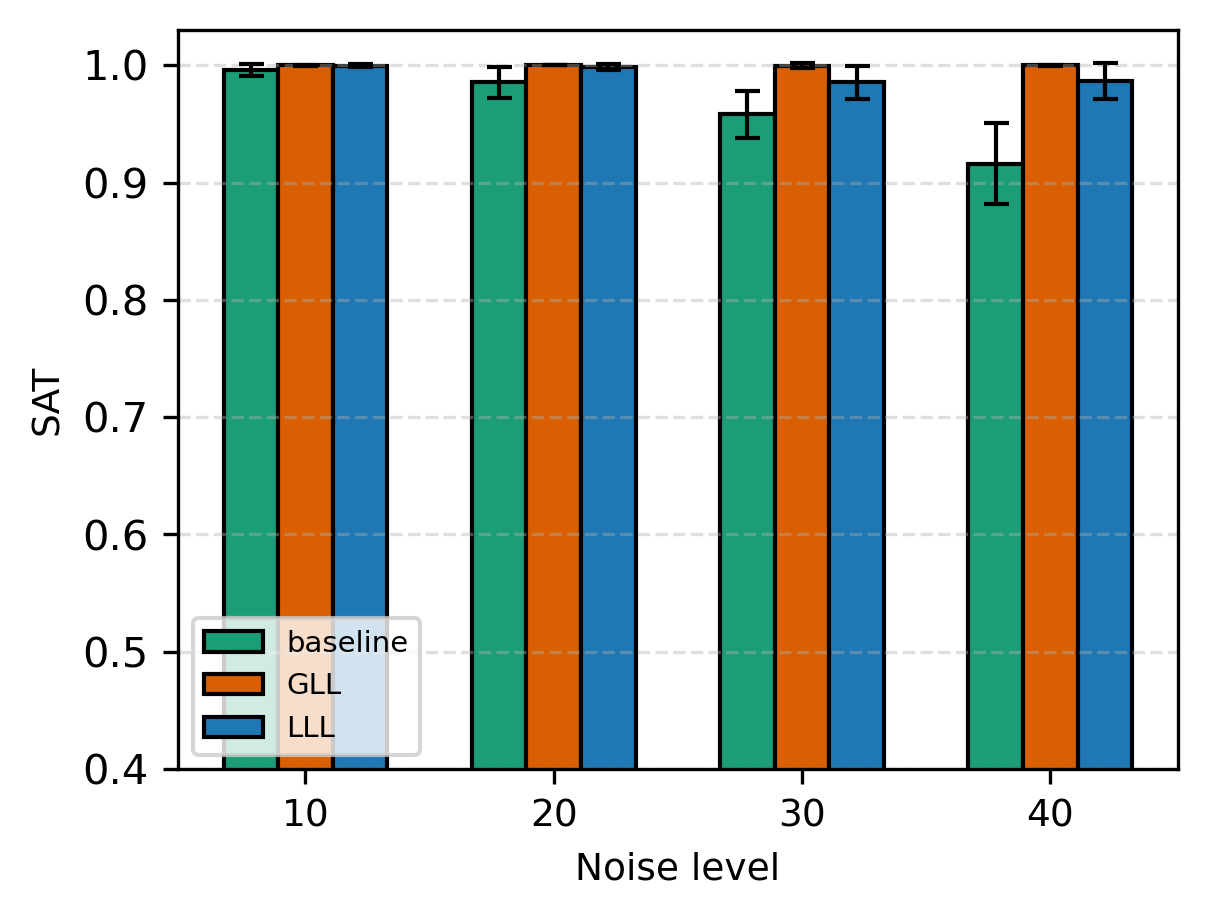}
  \end{subfigure}

  % BPIC 2020
  \vspace{0.3em}\noindent\textbf{BPIC 2020}\par\vspace{0.3em}
  \begin{subfigure}[b]{0.45\textwidth}
    \includegraphics[width=\linewidth]{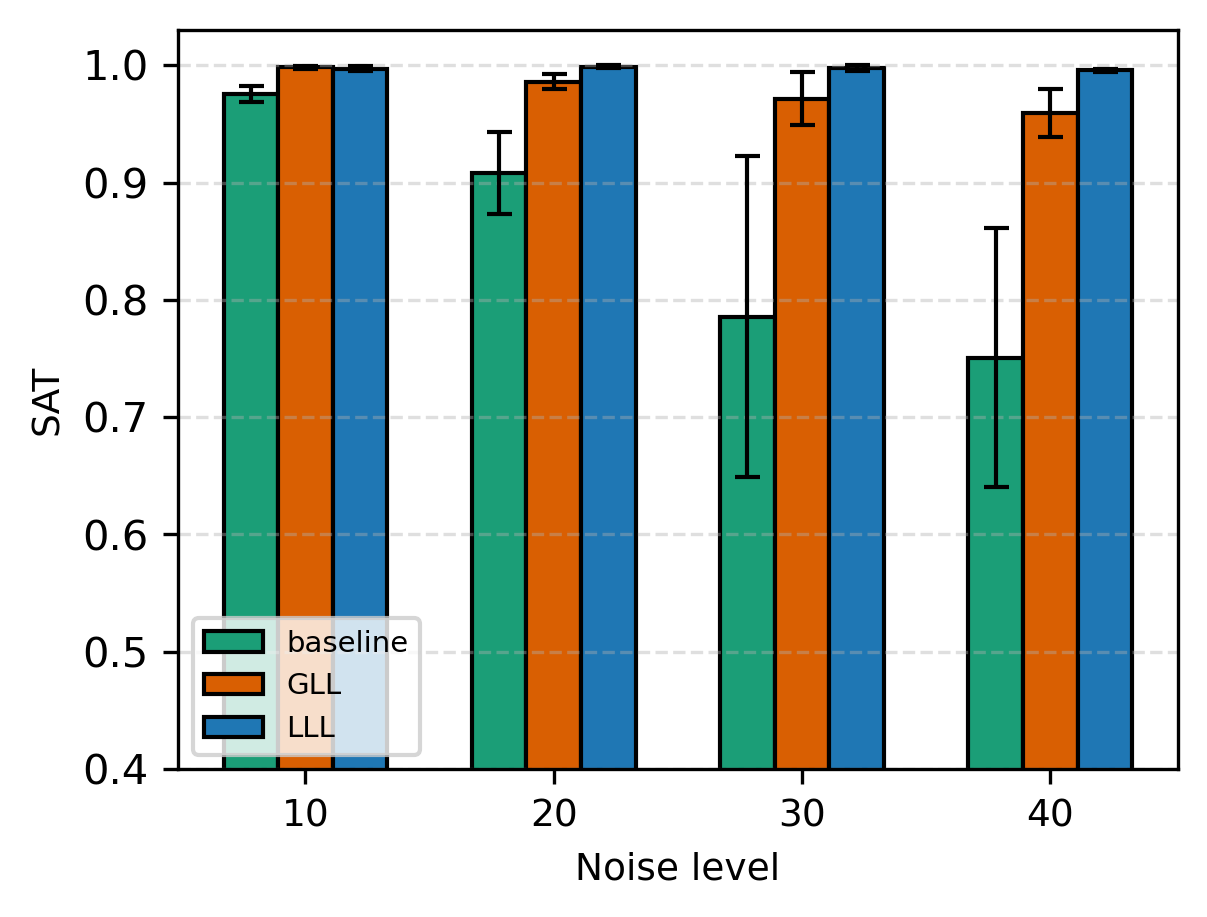}
  \end{subfigure}
  \begin{subfigure}[b]{0.45\textwidth}
    \includegraphics[width=\linewidth]{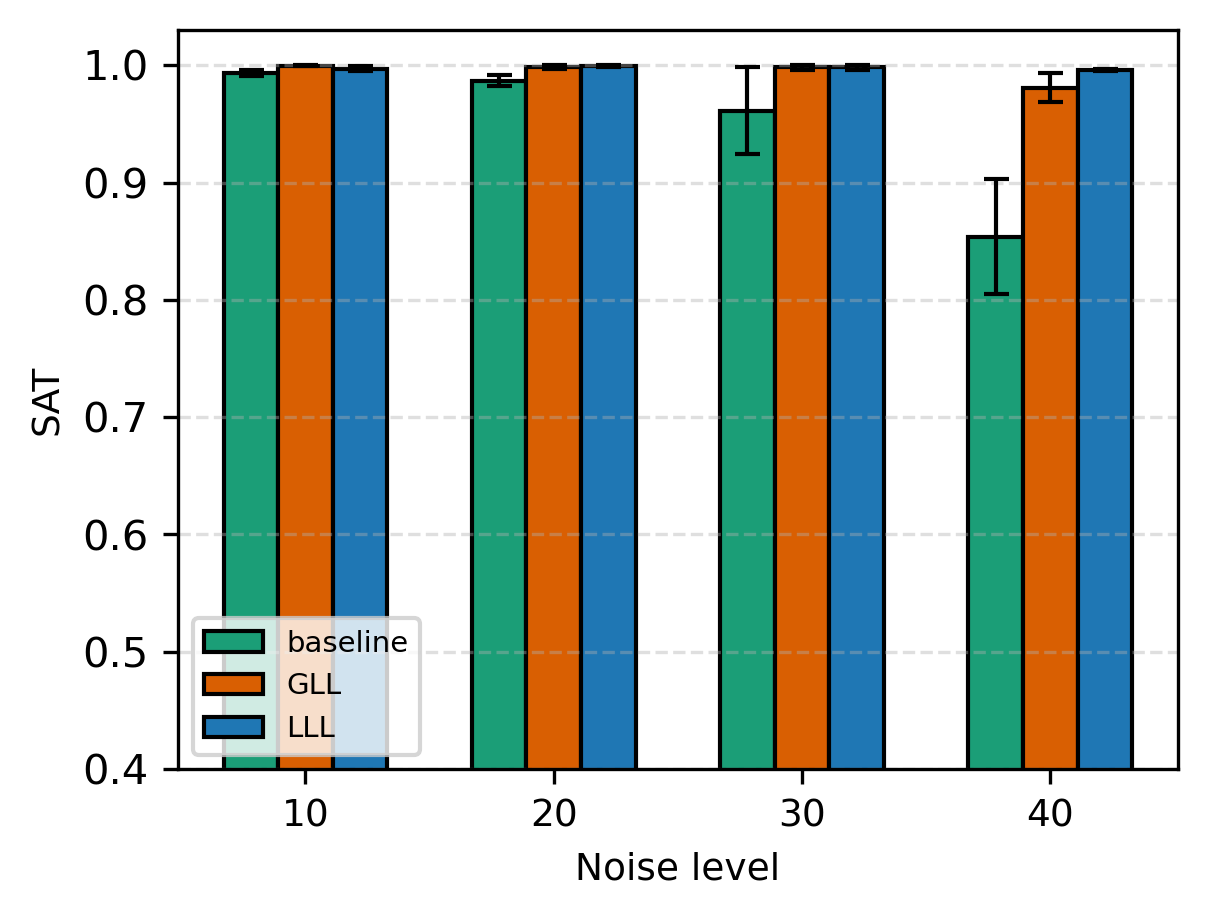}
  \end{subfigure}

  \caption{Satisfiability (SAT) of predicted traces under increasing noise levels for Transformer models. Columns correspond to the sampling strategies (temperature on the left, greedy on the right). Results are averaged over prefix lengths; error bars indicate the standard deviation over prefix lengths.}
  \label{fig:trans_sat_comparison}
\end{figure}

\begin{figure}[!htbp]
  \centering

  % Sepsis
  \vspace{0.3em}\noindent\textbf{Sepsis}\par\vspace{0.3em}
  \begin{subfigure}[b]{0.45\textwidth}
    \includegraphics[width=\linewidth]{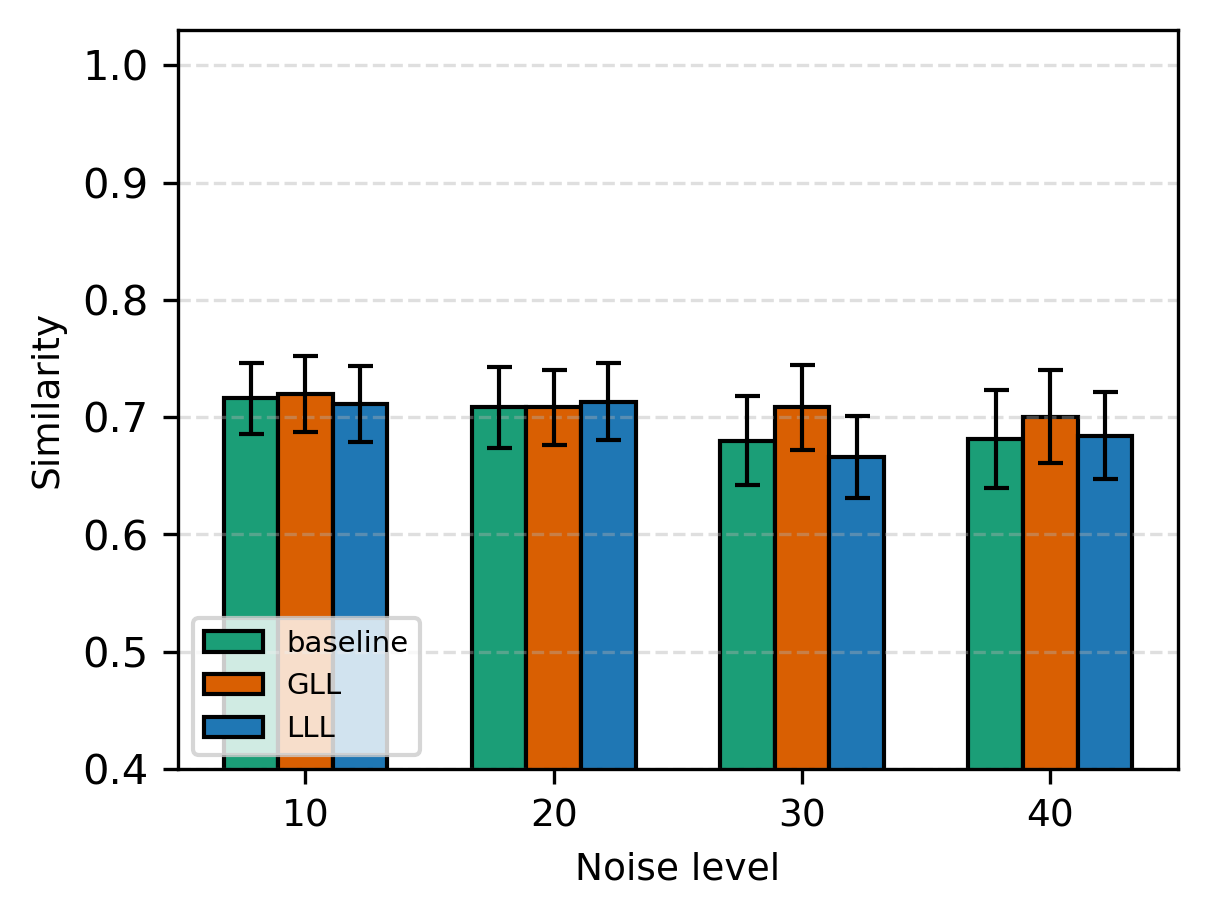}
  \end{subfigure}
  \begin{subfigure}[b]{0.45\textwidth}
    \includegraphics[width=\linewidth]{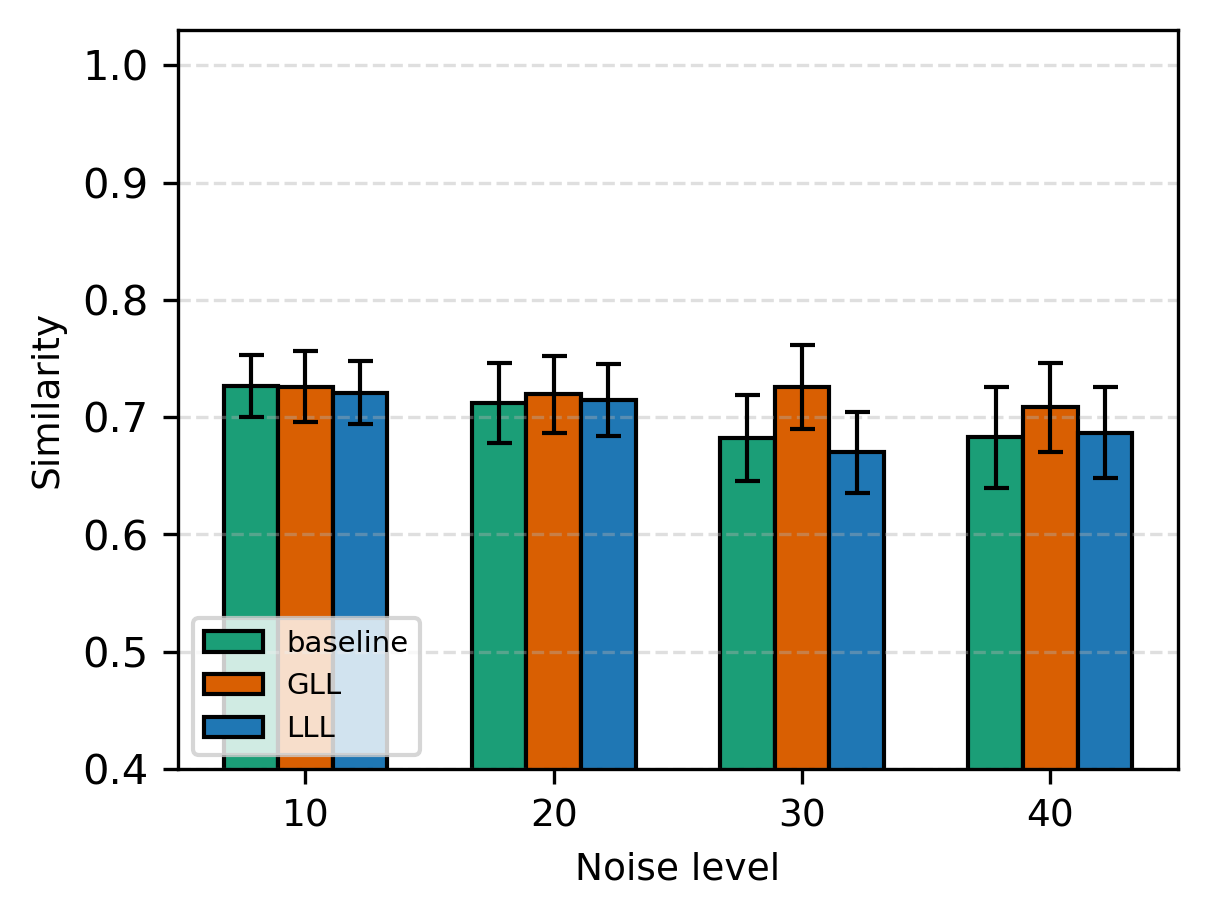}
  \end{subfigure}

  % BPIC 2013
  \vspace{0.3em}\noindent\textbf{BPIC 2013}\par\vspace{0.3em}
  \begin{subfigure}[b]{0.45\textwidth}
    \includegraphics[width=\linewidth]{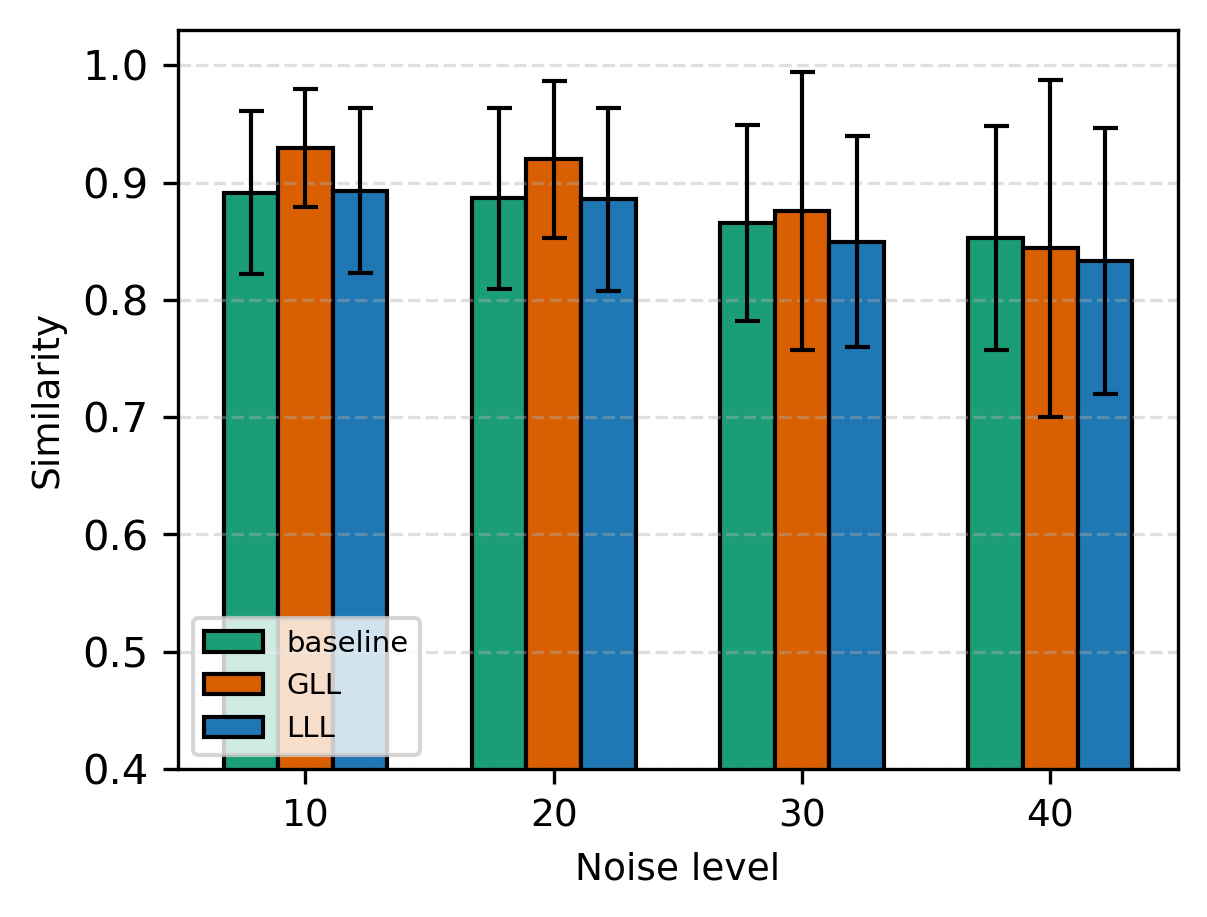}
  \end{subfigure}
  \begin{subfigure}[b]{0.45\textwidth}
    \includegraphics[width=\linewidth]{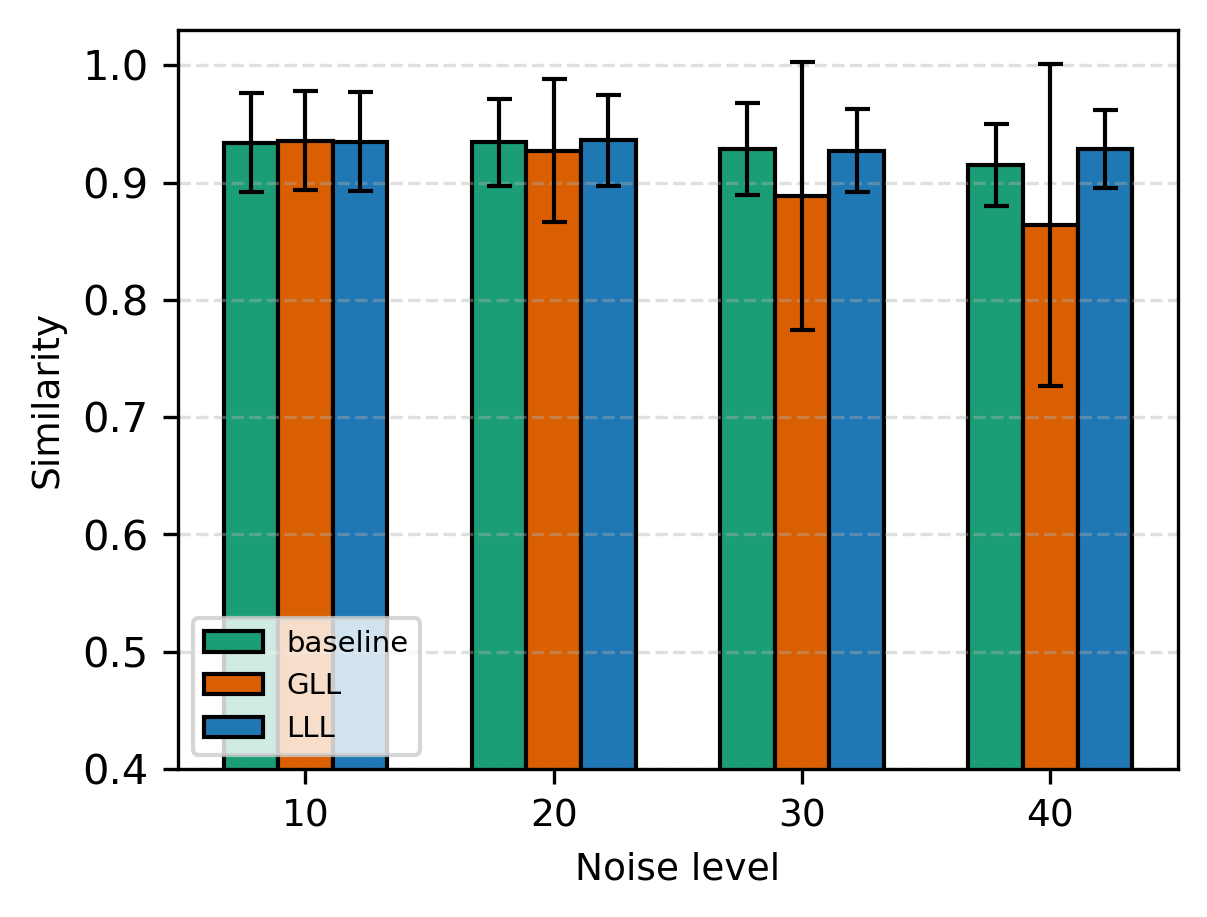}
  \end{subfigure}

  % BPIC 2020
  \vspace{0.3em}\noindent\textbf{BPIC 2020}\par\vspace{0.3em}
  \begin{subfigure}[b]{0.45\textwidth}
    \includegraphics[width=\linewidth]{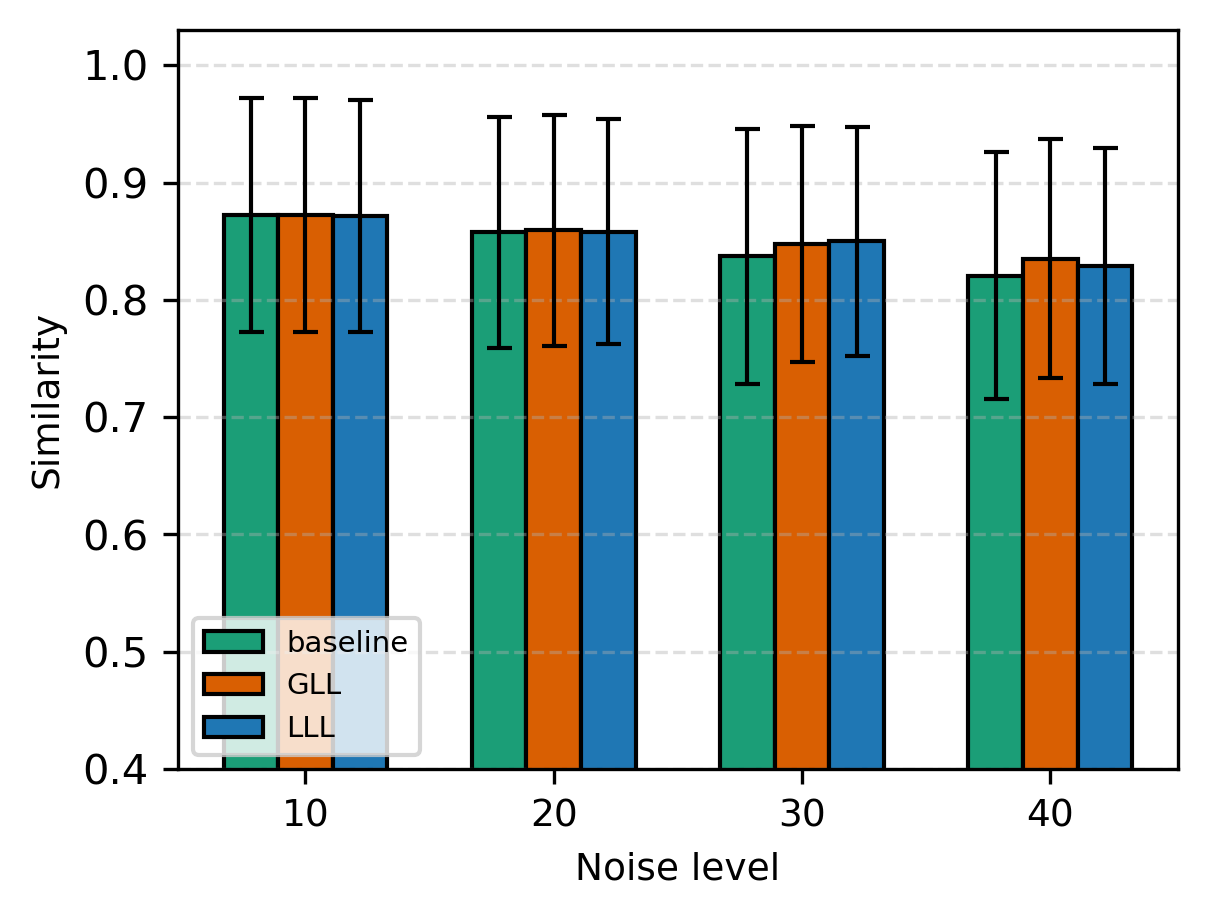}
  \end{subfigure}
  \begin{subfigure}[b]{0.45\textwidth}
    \includegraphics[width=\linewidth]{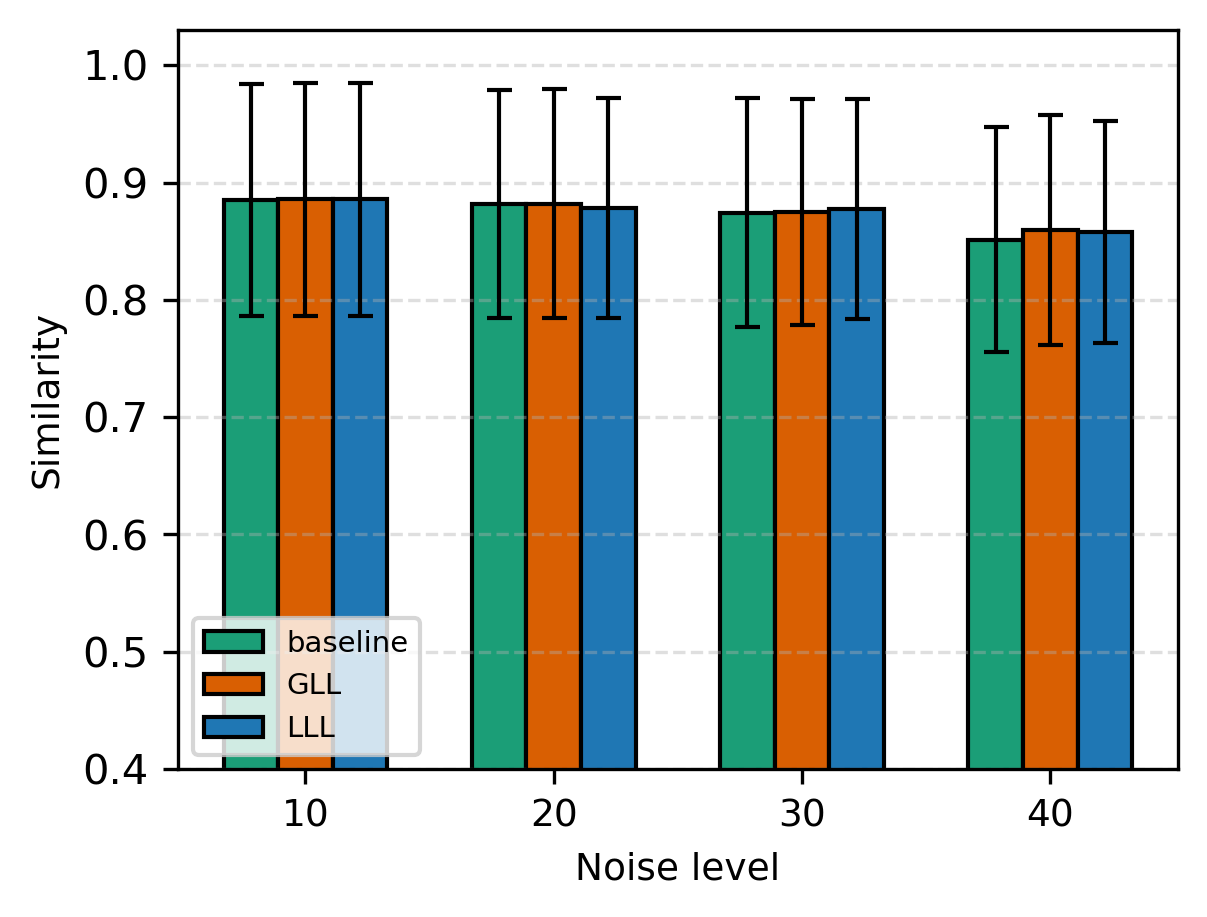}
  \end{subfigure}

  \caption{Normalized Damerau--Levenshtein similarity under increasing noise levels for Transformer models. Columns correspond to the sampling strategies (temperature on the left, greedy on the right). Results are averaged over prefix lengths; error bars indicate the standard deviation over prefix lengths.}
  \label{fig:trans_similarity_comparison}
\end{figure}

\textcolor{black}{
To further validate these findings, we replicate the same empirical evaluation using Transformer architectures (with $\alpha = 0.75$ for GLL and $\alpha = 0.25$ for LLL). Figures~\ref{fig:trans_sat_comparison} and~\ref{fig:trans_similarity_comparison} report the corresponding results and reveal behavior consistent with that observed for LSTMs. In particular, when either GLL or LLL is employed, the satisfiability remains close to 100\% across all noise levels, whereas in the baseline configuration it deteriorates substantially as noise increases. As with LSTMs, the improvement in satisfiability does not come at the expense of predictive quality: the suffix similarity remains comparable to the baseline across configurations. These results confirm that the benefits of incorporating the logic-based loss generalize across architectures and are not specific to recurrent models.
}

Differences observed across datasets likely reflect their inherent characteristics, such as vocabulary size, number and frequency of variants, and training data volume. Overall, our empirical evaluation confirms that integrating background knowledge at training time consistently improves performance and that the methodology is applicable across various real-world datasets and scenarios.
\textcolor{black}{Across all datasets and noise levels, the most important result is the clear improvement obtained by introducing the logic-based loss compared to the baseline. This result depends on a fixed value of the $\alpha$ hyperparameter in the combined loss function in Equation~\ref{eq:loss-balance}.}
\begin{table}[!h]
\color{black}
\centering
\scalebox{0.63}{
%\begin{tabular}{cc|cccccccccc|c}
%\toprule
%Dataset & Noise & 0.05 & 0.15 & 0.25 & 0.35 & 0.45 & 0.55 & 0.65 & 0.75 & 0.85 & 0.95 & 1 \\
%\midrule
%\multicolumn{2}{c|}{} & \multicolumn{10}{c|}{Global Guidance} & Baseline \\
\begin{tabular}{lc cccccccccc c}
\toprule
\multirow{2}{*}{Dataset} & \multirow{2}{*}{Noise} & \multicolumn{10}{c}{Global Guidance} & Baseline \\
\cmidrule(lr){3-12}\cmidrule(lr){13-13}
& & 0.05 & 0.15 & 0.25 & 0.35 & 0.45 & 0.55 & 0.65 & 0.75 & 0.85 & 0.95 & 1 \\
\midrule
\multirow{4}{*}{Sepsis} 
 & 10\% & \textbf{100\%} & \textbf{100\%} & \textbf{100\%} & \textbf{100\%} & \textbf{100\%} & \textbf{100\%} & 99.7\% & \textbf{100\%} & 99.8\% & 98.5\% & 85.0\% \\
 & 20\% & \textbf{100\%} & \textbf{100\%} & \textbf{100\%} & \textbf{100\%} & \textbf{100\%} & \textbf{100\%} & 99.8\% & \textbf{100\%} & 99.8\% & 98.2\% & 87.1\% \\
 & 30\% & \textbf{100\%} & \textbf{100\%} & \textbf{100\%} & \textbf{100\%} & \textbf{100\%} & \textbf{100\%} & 99.9\% & 99.6\% & 99.6\% & 96.8\% & 75.1\% \\
 & 40\% & \textbf{100\%} & \textbf{100\%} & \textbf{100\%} & 99.8\% & 99.9\% & 99.5\% & 99.8\% & 99.8\% & 99.3\% & 92.4\% & 58.9\% \\
\midrule
\multirow{4}{*}{BPIC13}
 & 10\% & \textbf{100\%} & \textbf{100\%} & \textbf{100\%} & \textbf{100\%} & \textbf{100\%} & \textbf{100\%} & \textbf{100\%} & \textbf{100\%} & \textbf{100\%} & \textbf{100\%} & 99.8\% \\
 & 20\% & \textbf{100\%} & \textbf{100\%} & \textbf{100\%} & \textbf{100\%} & \textbf{100\%} & \textbf{100\%} & \textbf{100\%} & \textbf{100\%} & \textbf{100\%} & 99.5\% & 99.0\% \\
 & 30\% & \textbf{100\%} & \textbf{100\%} & \textbf{100\%} & \textbf{100\%} & \textbf{100\%} & \textbf{100\%} & \textbf{100\%} & \textbf{100\%} & \textbf{100\%} & 99.6\% & 98.0\% \\
 & 40\% & \textbf{100\%} & \textbf{100\%} & \textbf{100\%} & \textbf{100\%} & \textbf{100\%} & \textbf{100\%} & \textbf{100\%} & \textbf{100\%} & \textbf{100\%} & 99.6\% & 90.5\% \\
\midrule
\multirow{4}{*}{BPIC20}
 & 10\% & \textbf{100\%} & \textbf{100\%} & \textbf{100\%} & \textbf{100\%} & \textbf{100\%} & \textbf{100\%} & \textbf{100\%} & \textbf{100\%} & \textbf{100\%} & 99.9\% & 99.6\% \\
 & 20\% & \textbf{100\%} & \textbf{100\%} & \textbf{100\%} & \textbf{100\%} & \textbf{100\%} & \textbf{100\%} & \textbf{100\%} & \textbf{100\%} & \textbf{100\%} & 99.9\% & 99.4\% \\
 & 30\% & \textbf{100\%} & \textbf{100\%} & \textbf{100\%} & \textbf{100\%} & \textbf{100\%} & \textbf{100\%} & \textbf{100\%} & \textbf{100\%} & 99.9\% & 99.5\% & 99.1\% \\
 & 40\% & \textbf{100\%} & \textbf{100\%} & \textbf{100\%} & \textbf{100\%} & \textbf{100\%} & \textbf{100\%} & \textbf{100\%} & \textbf{100\%} & 99.9\% & 88.6\% & 79.6\% \\
%\midrule
%\multicolumn{2}{c|}{} & \multicolumn{10}{c|}{Local Guidance} & Baseline \\
\bottomrule
\\
& & \multicolumn{10}{c}{Local Guidance} & Baseline \\
\cmidrule(lr){3-12}\cmidrule(lr){13-13}
& & 0.05 & 0.15 & 0.25 & 0.35 & 0.45 & 0.55 & 0.65 & 0.75 & 0.85 & 0.95 & 1 \\
\midrule
\multirow{4}{*}{Sepsis} 
 & 10\% & \textbf{100\%} & \textbf{100\%} & \textbf{100\%} & \textbf{100\%} & \textbf{100\%} & \textbf{100\%} & \textbf{100\%} & \textbf{100\%} & \textbf{100\%} & \textbf{100\%} & 85.0\% \\
 & 20\% & \textbf{100\%} & 99.9\% & 99.9\% & 99.9\% & 99.9\% & 99.9\% & 99.9\% & 99.9\% & 99.9\% & 99.9\% & 87.1\% \\
 & 30\% & 99.9\% & \textbf{100\%} & 99.9\% & 99.9\% & \textbf{100\%} & \textbf{100\%} & \textbf{100\%} & 99.9\% & \textbf{100\%} & 99.9\% & 75.1\% \\
 & 40\% & 98.6\% & 98.2\% & \textbf{98.8\%} & 98.7\% & 97.8\% & 97.9\% & \textbf{98.8\%} & \textbf{98.8\%} & 98.4\% & \textbf{98.8\%} & 58.9\% \\
\midrule
\multirow{4}{*}{BPIC13}
 & 10\% & \textbf{100\%} & \textbf{100\%} & \textbf{100\%} & \textbf{100\%} & \textbf{100\%} & \textbf{100\%} & \textbf{100\%} & \textbf{100\%} & \textbf{100\%} & \textbf{100\%} & 99.8\% \\
 & 20\% & \textbf{99.7\%} & \textbf{99.7\%} & \textbf{99.7\%} & \textbf{99.7\%} & \textbf{99.7\%} & \textbf{99.7\%} & \textbf{99.7\%} & \textbf{99.7\%} & \textbf{99.7\%} & \textbf{99.7\%} & 99.0\% \\
 & 30\% & 99.3\% & \textbf{99.5\%} & \textbf{99.5\%} & 99.4\% & 99.4\% & 99.4\% & 99.4\% & 99.4\% & \textbf{99.5\%} & 99.4\% & 98.0\% \\
 & 40\% & 99.3\% & 99.3\% & 99.1\% & 99.2\% & 99.2\% & 98.9\% & \textbf{99.4\%} & 99.2\% & 99.2\% & 99.2\% & 90.5\% \\
\midrule
\multirow{4}{*}{BPIC20}
 & 10\% & \textbf{99.9\%} & 99.8\% & 99.8\% & 99.8\% & \textbf{99.9\%} & 99.8\% & \textbf{99.9\%} & \textbf{99.9\%} & 99.8\% & \textbf{99.9\%} & 99.6\% \\
 & 20\% & \textbf{100\%} & \textbf{100\%} & \textbf{100\%} & \textbf{100\%} & 99.9\% & 99.9\% & 99.9\% & \textbf{100\%} & 99.9\% & 99.9\% & 99.4\% \\
 & 30\% & \textbf{100\%} & \textbf{100\%} & \textbf{100\%} & \textbf{100\%} & \textbf{100\%} & \textbf{100\%} & \textbf{100\%} & \textbf{100\%} & 99.9\% & 99.9\% & 99.1\% \\
 & 40\% & \textbf{99.6\%} & \textbf{99.6\%} & 99.5\% & 99.5\% & 99.5\% & 99.5\% & 99.5\% & 99.5\% & 99.5\% & 99.5\% & 79.6\% \\
\bottomrule
\end{tabular}
}
\caption{Satisfiability scores obtained with the LSTM model for different values of the parameter $\alpha$ using greedy sampling. The rightmost column reports the baseline setting ($\alpha = 1$), in which no logic-based loss is used. The best results are in bold.}
\label{tab:satisfiability_alpha}
\end{table}
\begin{table}[!t]
\color{black}
\centering
\scalebox{0.63}{
%\begin{tabular}{cc|cccccccccc|c}
%\toprule
%Dataset & Noise & 0.05 & 0.15 & 0.25 & 0.35 & 0.45 & 0.55 & 0.65 & 0.75 & 0.85 & 0.95 & 1 \\
%\midrule
%\multicolumn{2}{c|}{} & \multicolumn{10}{c|}{Global Guidance} & Baseline \\
\begin{tabular}{lc cccccccccc c}
\toprule
\multirow{2}{*}{Dataset} & \multirow{2}{*}{Noise} & \multicolumn{10}{c}{Global Guidance} & Baseline \\
\cmidrule(lr){3-12}\cmidrule(lr){13-13}
& & 0.05 & 0.15 & 0.25 & 0.35 & 0.45 & 0.55 & 0.65 & 0.75 & 0.85 & 0.95 & 1 \\
\midrule
\multirow{4}{*}{Sepsis} 
 & 10\% & 52.2\% & 52.9\% & 57.2\% & 61.9\% & 68.6\% & 69.9\% & 70.8\% & 72.9\% & 72.9\% & 73.1\% & \textbf{73.9\%} \\
 & 20\% & 49.5\% & 50.8\% & 63.3\% & 65.3\% & 72.7\% & 74.1\% & 73.1\% & \textbf{74.6\%} & 71.9\% & 73.4\% & 72.5\% \\
 & 30\% & 47.6\% & 49.2\% & 54.9\% & 65.9\% & 68.7\% & 71.6\% & 72.3\% & \textbf{74.5\%} & 72.2\% & 68.2\% & 68.1\% \\
 & 40\% & 55.5\% & 55.6\% & 55.4\% & 66.0\% & 69.0\% & 70.3\% & 72.2\% & 72.6\% & \textbf{73.8\%} & 69.9\% & 68.1\% \\
\midrule
\multirow{4}{*}{BPIC13}
 & 10\% & 81.3\% & 81.4\% & 81.5\% & 81.6\% & 81.5\% & 82.7\% & 86.8\% & \textbf{93.9\%} & \textbf{93.9\%} & \textbf{93.9\%} & 93.3\% \\
 & 20\% & 81.2\% & 81.5\% & 81.5\% & 81.4\% & 81.4\% & 81.4\% & 81.3\% & \textbf{93.9\%} & \textbf{93.9\%} & 93.7\% & 93.5\% \\
 & 30\% & 81.2\% & 81.3\% & 81.2\% & 81.2\% & 81.2\% & 81.2\% & 81.2\% & 88.1\% & \textbf{93.9\%} & 93.7\% & 93.5\% \\
 & 40\% & 81.2\% & 81.3\% & 81.3\% & 81.4\% & 81.2\% & 81.3\% & 81.2\% & 81.9\% & \textbf{93.2\%} & 93.1\% & 91.1\% \\
\midrule
\multirow{4}{*}{BPIC20}
 & 10\% & 88.4\% & 88.5\% & \textbf{88.6\%} & \textbf{88.6\%} & \textbf{88.6\%} & \textbf{88.6\%} & \textbf{88.6\%} & \textbf{88.6\%} & \textbf{88.6\%} & 88.5\% & \textbf{88.6\%} \\
 & 20\% & 88.3\% & 88.4\% & \textbf{88.5\%} & \textbf{88.5\%} & \textbf{88.5\%} & \textbf{88.5\%} & \textbf{88.5\%} & \textbf{88.5\%} & \textbf{88.5\%} & 88.4\% & 88.4\% \\
 & 30\% & 88.4\% & \textbf{88.5\%} & \textbf{88.5\%} & \textbf{88.5\%} & \textbf{88.5\%} & \textbf{88.5\%} & \textbf{88.5\%} & 88.4\% & 88.3\% & 88.2\% & 87.7\% \\
 & 40\% & 87.4\% & 88.3\% & \textbf{88.4\%} & 88.0\% & 87.7\% & 87.2\% & 87.6\% & 86.9\% & 86.8\% & 86.3\% & 85.9\% \\
%\midrule
%\multicolumn{2}{c|}{} & \multicolumn{10}{c|}{Local Guidance} & Baseline \\
\bottomrule
\\
& & \multicolumn{10}{c}{Local Guidance} & Baseline \\
\cmidrule(lr){3-12}\cmidrule(lr){13-13}
& & 0.05 & 0.15 & 0.25 & 0.35 & 0.45 & 0.55 & 0.65 & 0.75 & 0.85 & 0.95 & 1 \\
\midrule
\multirow{4}{*}{Sepsis} 
 & 10\% & 61.1\% & \textbf{73.9\%} & 73.2\% & 73.2\% & 73.1\% & 72.7\% & 72.7\% & 72.9\% & 72.6\% & 73.3\% & \textbf{73.9\%} \\
 & 20\% & 68.8\% & \textbf{73.2\%} & \textbf{73.2\%} & 73.0\% & 72.9\% & 72.7\% & 72.7\% & 72.6\% & 72.6\% & 72.5\% & 72.5\% \\
 & 30\% & \textbf{73.3\%} & 70.6\% & 71.0\% & 69.2\% & 70.3\% & 70.6\% & 70.6\% & 70.8\% & 70.4\% & 70.9\% & 68.1\% \\
 & 40\% & 71.1\% & 71.8\% & \textbf{72.8\%} & 72.5\% & \textbf{72.8\%} & 72.7\% & 72.7\% & 72.7\% & 72.5\% & 72.7\% & 68.1\% \\
\midrule
\multirow{4}{*}{BPIC13}
 & 10\% & 93.7\% & 93.7\% & \textbf{93.8\%} & \textbf{93.8\%} & \textbf{93.8\%} & \textbf{93.8\%} & 93.6\% & 93.7\% & 93.6\% & 93.4\% & 93.3\% \\
 & 20\% & \textbf{93.6\%} & \textbf{93.6\%} & \textbf{93.6\%} & \textbf{93.6\%} & \textbf{93.6\%} & \textbf{93.6\%} & \textbf{93.6\%} & \textbf{93.6\%} & \textbf{93.6\%} & \textbf{93.6\%} & 93.5\% \\
 & 30\% & 93.3\% & 93.4\% & 93.4\% & 93.3\% & \textbf{93.5\%} & 93.4\% & 93.2\% & 93.2\% & \textbf{93.5\%} & \textbf{93.5\%} & \textbf{93.5\%} \\
 & 40\% & \textbf{93.3\%} & \textbf{93.3\%} & 93.2\% & 93.2\% & \textbf{93.3\%} & 93.1\% & 93.2\% & 93.2\% & 93.2\% & 93.2\% & 91.1\% \\
\midrule
\multirow{4}{*}{BPIC20}
 & 10\% & 88.5\% & \textbf{88.7\%} & 88.6\% & 88.6\% & 88.6\% & 88.6\% & 88.6\% & 88.6\% & 88.6\% & 88.6\% & 88.6\% \\
 & 20\% & 88.3\% & 88.3\% & \textbf{88.4\%} & 88.3\% & \textbf{88.4\%} & 88.3\% & \textbf{88.4\%} & \textbf{88.4\%} & \textbf{88.4\%} & \textbf{88.4\%} & \textbf{88.4\%} \\
 & 30\% & \textbf{88.3\%} & 87.8\% & 87.8\% & 87.7\% & 87.7\% & 87.6\% & 87.6\% & 87.6\% & 87.5\% & 87.5\% & 87.7\% \\
 & 40\% & \textbf{87.8\%} & 86.9\% & 86.5\% & 86.6\% & 87.1\% & 86.9\% & 86.8\% & 86.9\% & 87.0\% & 87.2\% & 85.9\% \\
\bottomrule
\end{tabular}
}
\caption{Similarity scores obtained with the LSTM model for different values of the parameter $\alpha$ using greedy sampling. The rightmost column reports the baseline setting ($\alpha = 1$), in which no logic-based loss is used. The best results are in bold.}
\label{tab:similarity_alpha}
\end{table}
\textcolor{black}{
\\ \indent We now explore the impact of this hyperparameter by showing how performance differs across different values of $\alpha$. As shown in Table~\ref{tab:satisfiability_alpha}, satisfiability in the baseline setting decreases noticeably as noise increases, while the inclusion of the logic loss keeps satisfiability close to perfect over a broad range of $\alpha$ values. This demonstrates that the logic component is crucial for ensuring that generated outputs satisfy the required constraints, especially under noisy conditions. For similarity, the effect of the logic loss depends on the guidance strategy, as shown in Table~\ref{tab:similarity_alpha}. With Global Guidance, giving more weight to the logic loss can sometimes reduce similarity, indicating a trade-off between strictly enforcing constraints and staying close to the original sequences. However, Global Guidance still achieves the best similarity scores in many settings when $\alpha$ is properly chosen, showing that strong logical guidance can also align well with task performance. Local Guidance, instead, provides more stable similarity across different $\alpha$ values and noise levels, while still benefiting from the large improvements in satisfiability. Overall, the logic-based loss is essential for improving constraint satisfaction over the baseline; Global Guidance often reaches the highest peak performance, and Local Guidance offers a more consistent balance between logical correctness and similarity.
}
\begin{table}[!h]
\color{black}
\centering
\scalebox{0.65}{
%\begin{tabular}{cc|cc|cc|c|cc|cc|c}
%\toprule
%Dataset & Noise & \multicolumn{5}{c|}{Satisfiability} & \multicolumn{5}{c}{Similarity} \\
%\midrule
% & & \multicolumn{2}{c|}{GLL} & \multicolumn{2}{c|}{LLL} & Baseline & \multicolumn{2}{c|}{GLL} & \multicolumn{2}{c|}{LLL} & Baseline \\
% & & 0.25 & 0.75 & 0.25 & 0.75 & 1 & 0.25 & 0.75 & 0.25 & 0.75 & 1 \\
\begin{tabular}{lc cc cc c cc cc c}
\toprule
\multirow{3}{*}{Dataset} & \multirow{3}{*}{Noise} & \multicolumn{5}{c}{Satisfiability} & \multicolumn{5}{c}{Similarity} \\
\cmidrule(lr){3-7}\cmidrule(lr){8-12}
 & & \multicolumn{2}{c}{GLL} & \multicolumn{2}{c}{LLL} & Baseline & \multicolumn{2}{c}{GLL} & \multicolumn{2}{c}{LLL} & Baseline \\
\cmidrule(lr){3-4}\cmidrule(lr){5-6}\cmidrule(lr){7-7}\cmidrule(lr){8-9}\cmidrule(lr){10-11}\cmidrule(lr){12-12}
 & & 0.25 & 0.75 & 0.25 & 0.75 & 1 & 0.25 & 0.75 & 0.25 & 0.75 & 1 \\
\midrule
\multirow{4}{*}{Sepsis} 
 & 10\% & \textbf{99.9\%} & 98.6\% & \textbf{99.9\%} & \textbf{99.9\%} & 90.1\% & \textbf{73.1\%} & 72.6\% & 72.1\% & 72.0\% & 72.6\% \\
 & 20\% & \textbf{99.9\%} & 98.7\% & \textbf{99.9\%} & \textbf{99.9\%} & 83.4\% & 71.1\% & 71.9\% & 71.5\% & \textbf{72.0\%} & 71.2\% \\
 & 30\% & 99.7\% & 97.3\% & 99.9\% & \textbf{100\%} & 69.2\% & 70.1\% & \textbf{72.6\%} & 67.0\% & 69.5\% & 68.2\% \\
 & 40\% & \textbf{99.1\%} & 95.1\% & 98.9\% & 98.6\% & 63.2\% & 69.5\% & \textbf{70.8\%} & 68.7\% & 69.5\% & 68.3\% \\
\midrule
\multirow{4}{*}{BPIC 2013}
 & 10\% & \textbf{100\%} & \textbf{100\%} & \textbf{100\%} & 99.9\% & 99.6\% & 81.7\% & \textbf{93.6\%} & 93.5\% & 93.5\% & 93.4\% \\
 & 20\% & \textbf{100\%} & \textbf{100\%} & 99.8\% & 99.7\% & 98.5\% & 81.5\% & 92.7\% & \textbf{93.6\%} & 93.5\% & 93.4\% \\
 & 30\% & \textbf{100\%} & \textbf{100\%} & 98.5\% & 98.9\% & 95.8\% & 81.2\% & 88.9\% & 92.7\% & \textbf{93.0\%} & 92.9\% \\
 & 40\% & \textbf{100\%} & \textbf{100\%} & 98.6\% & 98.6\% & 91.6\% & 81.2\% & 86.4\% & \textbf{92.9\%} & \textbf{92.9\%} & 91.5\% \\
\midrule
\multirow{4}{*}{BPIC 2020}
 & 10\% & \textbf{100\%} & \textbf{100\%} & 99.7\% & 99.7\% & 99.4\% & 88.3\% & \textbf{88.6\%} & \textbf{88.6\%} & \textbf{88.6\%} & 88.5\% \\
 & 20\% & \textbf{100\%} & 99.9\% & 99.9\% & 99.9\% & 98.7\% & \textbf{88.3\%} & 88.2\% & 87.8\% & 88.2\% & 88.2\% \\
 & 30\% & \textbf{100\%} & 99.8\% & 99.8\% & 99.9\% & 96.1\% & \textbf{87.9\%} & 87.5\% & 87.7\% & \textbf{87.9\%} & 87.4\% \\
 & 40\% & \textbf{99.9\%} & 98.1\% & 99.6\% & 99.6\% & 85.4\% & \textbf{86.5\%} & 85.9\% & 85.8\% & 86.1\% & 85.1\% \\
\bottomrule
\end{tabular}
}
\caption{Satisfiability and similarity scores obtained with the Transformer model for different values of the parameter $\alpha$ using greedy sampling. The rightmost column corresponds to the baseline setting ($\alpha = 1$), in which no logic-based loss is used. The best results are in bold.}
\label{tab:gll_sim_alpha}
\end{table}

\textcolor{black}{Experiments with Transformers corroborate the previously observed trends, as shown in Table~\ref{tab:gll_sim_alpha}, which reports both satisfiability and similarity for $\alpha \in \{0.25, 0.75\}$, together with the baseline. As expected, increasing the noise level in the training data consistently degrades baseline performance. In terms of satisfiability, both Global and Local Guidance keep constraint satisfaction near 100\% across all noise levels, with Global Guidance providing the strongest results. Similarity again exhibits more varied behavior: Local Guidance yields higher similarity in some configurations, whereas Global Guidance is generally more robust under high noise. Notably, the baseline is uniformly worse across settings, underscoring the relevance of the logical component and the overall robustness of the proposed approach.}

\textcolor{black}{We also report the average number of training epochs in Table~\ref{tab:avg_epochs}. We observe that incorporating the logic-based loss consistently reduces the number of epochs required for convergence across all datasets, noise levels, and neural architectures, with the effect being particularly pronounced for the LSTM models. This suggests that logic-guided objectives improve optimization efficiency in terms of training iterations. For LSTM models, GLL generally requires fewer epochs than LLL, in particular for the datasets (Sepsis and BPIC 2020) that have low compliance with the background knowledge (see Table~\ref{tab:dataset_acceptance}). Indeed, the maximum difference in training epochs is around 310 for both datasets, whereas for BPIC 2013 the maximum difference is around 25 training epochs. This suggests that, for LSTM models, GLL can be preferred for datasets with high inconsistency with the background knowledge.}
%\textcolor{black}{When considering total training time, a more nuanced picture emerges. For RNN models, the reduction in epochs typically translates into lower overall training time, in line with the observed convergence improvements. In contrast, for Transformers the baseline often achieves lower total training time. This behavior is attributable to the additional computational overhead introduced by the logic-based loss, which increases the per-epoch cost. In high-capacity architectures such as Transformers, this overhead can outweigh the gains obtained from faster convergence in terms of iterations.}
\begin{table}[!h]
\color{black}
\centering
\scalebox{0.85}{
\begin{tabular}{l c  c c c  c c c}
\toprule
\multirow{2}{*}{Dataset} & \multirow{2}{*}{Noise} & \multicolumn{3}{c}{LSTM} & \multicolumn{3}{c}{Transformer} \\ 
\cmidrule(lr){3-5}\cmidrule(lr){6-8}
 & & Baseline & GLL & LLL & Baseline & GLL & LLL \\
\midrule
\multirow{4}{*}{Sepsis}
 & 10\% & 1327.13 & \textbf{568.37} & 880.07 & 667.33 & \textbf{559.87} & 571.73 \\
 & 20\% & 1460.93 & \textbf{578.07} & 785.13 & 615.20 & 571.00 & \textbf{562.87} \\
 & 30\% & 1465.93 & \textbf{578.17} & 801.23 & 646.33 & \textbf{563.20} & 575.07 \\
 & 40\% & 1606.60 & \textbf{589.80} & 657.40 & 641.13 & 570.20 & \textbf{563.07} \\
\midrule
\multirow{4}{*}{BPIC 2013}
 & 10\% & 1477.73 & \textbf{562.93} & 587.60 & 571.07 & 563.47 & \textbf{561.20} \\
 & 20\% & 1565.93 & \textbf{566.80} & 587.80 & 568.53 & 566.80 & \textbf{558.13} \\
 & 30\% & 1895.73 & \textbf{572.00} & 583.77 & 571.47 & 572.77 & \textbf{564.70} \\
 & 40\% & 1998.80 & \textbf{575.90} & 601.20 & 571.33 & 578.63 & \textbf{564.20} \\
\midrule
\multirow{4}{*}{BPIC 2020}
 & 10\% & 1962.20 & \textbf{556.13} & 598.17 & 620.73 & \textbf{553.07} & 574.63 \\
 & 20\% & 1843.73 & \textbf{570.10} & 618.03 & 641.33 & \textbf{564.07} & 573.83 \\
 & 30\% & 1943.20 & \textbf{568.27} & 681.23 & 664.40 & \textbf{564.07} & 568.83 \\
 & 40\% & 1999.00 & \textbf{578.40} & 884.57 & 770.40 & \textbf{566.87} & 576.77 \\
\bottomrule
\end{tabular}
}
\caption{Average number of training epochs (the lower the better) required by each method across datasets, noise levels and neural architectures. Best results are in bold.}
\label{tab:avg_epochs}
\end{table}
\indent \textcolor{black}{Transformers generally have the effect of lowering the number of training epochs. This is not surprising because the self-attention mechanism provides direct connections between all events in a prefix, enabling faster and more effective credit assignment than the recurrent propagation through time performed by LSTM models. Here, credit assignment refers to the identification of which past inputs are responsible for the current prediction error, and therefore which parameters should be updated to correct it. In addition, the absence of recurrence, aided by residual connections and layer normalization, avoids vanishing or noisy gradients. Moreover, Transformers process the entire prefix in parallel, rather than sequentially as in LSTM models, thus allowing the network to quickly identify the most relevant past activities for predicting the future suffix. As a consequence, meaningful dependencies are learned earlier in training, leading to a lower number of training epochs. For this architecture, the difference in terms of training epochs between GLL and LLL is much more contained.
}

\section{Related Work}
\label{sec:related}
Recently, there has been significant interest in employing deep neural networks (NNs) in PPM for tasks such as next-activity prediction, suffix prediction, and attribute prediction~\cite{deeplearning_BPM_survey}. Despite significant advances in the field, nearly all works rely on training these models solely on data, without utilizing any formal prior knowledge about the process. They mainly focus on two aspects: (i) enhancing the neural model, ranging from RNNs~\cite{Tax17,9_,DiFrancescomarino17,11_} and convolutional neural networks (CNNs)~\cite{13_} to generative adversarial networks (GANs)~\cite{22_,11_}, autoencoders~\cite{11_}, and Transformers~\cite{11_}; and (ii) wisely choosing the sampling technique used to query the network at test time to generate the suffix, mostly using greedy search~\cite{Tax17}, random search~\cite{9_}, or beam search~\cite{DiFrancescomarino17}, and more recently, policies trained with reinforcement learning (RL)~\cite{6_,ramamaneiro24}.
Among all these works, only one exploits prior process knowledge~\cite{DiFrancescomarino17}, expressed as a set of \ltlf formulas, but it uses this knowledge only at test time, modifying the beam search sampling algorithm to select traces that are potentially compliant with the background knowledge.

In this work, we take a radically different approach by introducing a principled way to integrate background knowledge expressed in \ltlf into a deep NN model for suffix prediction at \textit{training time}. This is based on defining a logical loss that can be combined with the loss of any autoregressive neural model and used with any sampling technique at test time, drawing inspiration from the literature in Neuro-Symbolic AI~\cite{nesy_survey}. In this field, many prior works focus on exploiting temporal logical knowledge in deep learning tasks, but none have addressed multi-step symbolic sequence generation.

T-leaf~\cite{t-leaf} creates a semantic embedding space to represent both formulas and traces and uses it in tasks such as sequential action recognition and imitation learning, which do not involve multi-step prediction. In~\cite{umili_kr23}, an extension of Logic Tensor Networks (LTNs)~\cite{LTN,SerafiniGBDSB21} to represent fuzzy automata is proposed and employed to integrate \ltlf background knowledge into image sequence classification tasks. STLnet~\cite{STLnet} adopts a student–teacher training scheme in which the student network proposes a suffix based on the data, which is then corrected by the teacher network to satisfy the formula. This work uses Signal Temporal Logic (STL) formulas and is applied to continuous trajectories rather than discrete traces. Our attempts to apply it to discrete data and \ltlf formulas translated into STL yielded poor results, as the resulting STL formulas were extremely challenging for the framework to handle.

A recent line of research focuses on constraining Large Language Models (LLMs) with structured temporal knowledge, either by employing constrained beam search~\cite{trident,llm_beam_search_1,llm_beam_search_2,10.1145/3810944,abs-2312-08847}, training auxiliary models~\cite{llm_aux_mod_1,llm_aux_mod_2}, or exploiting conditioned sampling techniques~\cite{llm_sampling1,montecarlo_llm}. However, all these approaches are exclusively designed for test-time inference and have no influence on the training of the LLM. While this may be reasonable in the context of LLMs, where prior knowledge is often available only for specific subtasks, in PPM structured global knowledge about the process may be available \emph{before} data collection. In such cases, incorporating this knowledge during training, rather than only at inference time, can significantly benefit the learning process.

Our work is the first to integrate temporal knowledge into the generation of multi-step symbolic sequences at training time. It is based on encoding \ltlf formulas using a matrix representation that we have previously used for very different tasks, such as learning RL policies for non-Markovian tasks~\cite{umili_nesy_2023} and inducing automata from a set of labeled traces~\cite{deepdfa_ecai2024}, which we adapt here for use in the generative task of suffix prediction.

\section{Conclusions and Future Work}
This paper introduces a novel Neuro-Symbolic approach that seamlessly integrates temporal logic knowledge, expressed in \ltlf, into the training of neural suffix predictors for PPM. By combining data-driven learning with formal background knowledge, our approach achieves improved prediction accuracy and higher compliance with logical constraints, even under noisy conditions. The proposed logical loss formulations, offering both local and global perspectives, demonstrate the effectiveness and generality of the method across different real-world datasets.

Future work will focus on extending the logical loss framework to support additional types of constraints that capture diverse process dimensions, such as resources, numeric attributes, and event timestamps. We also aim to assess the method in the presence of concept drift, where process behavior evolves over time. Finally, further investigation into the synergy between local and global constraints, as well as their integration with Large Language Models, could prove promising for advancing the state-of-the-art in multi-step symbolic sequence generation.
\label{sec:conclusion}

\section*{Acknowledgments}
This work was carried out while Axel Mezini and Matteo Mancanelli were enrolled in the Italian National Doctorate on Artificial Intelligence run by Sapienza University of Rome in collaboration with the Free University of Bozen-Bolzano. The work of Fabio Patrizi, Elena Umili, and Matteo Mancanelli was supported by the PNRR MUR project PE0000013-FAIR. This study was funded by the European Union – NextGenerationEU, within the framework of the iNEST – Interconnected Nord-Est Innovation Ecosystem (iNEST ECS00000043 – CUP I43C22000250006). We also acknowledge financial support under the National Recovery and Resilience Plan (NRRP), M4C2I1.1, funded by the European Union – NextGenerationEU – Project Title: \textit{aRtificial intElligence for Process Analytics (REPA)} – Grant Assignment Decree No. 2022CJWPNA by the Italian Ministry of University and Research (MUR). The views and opinions expressed are solely those of the authors and do not necessarily reflect those of the European Union, nor can the European Union be held responsible for them.

\bibliographystyle{elsarticle-num} 
\bibliography{bibliography}

\end{document}